\RequirePackage[OT1]{fontenc}
\documentclass[letterpaper, 10 pt, conference]{ieeeconf}

\usepackage{ifthen}
\newcommand\arxivmode{true} 

\IEEEoverridecommandlockouts
\usepackage{graphicx}
\DeclareGraphicsExtensions{.pdf,.png,.jpg}

\usepackage{stfloats}
\usepackage{mwe}
\usepackage{bbm}
\usepackage{amssymb,amsopn,amstext,amsfonts,amsmath}

\let\proof\relax

\usepackage{amsthm}

\usepackage{balance}
\usepackage{xcolor}
\usepackage[bookmarks=true]{hyperref}
\usepackage{url}
\usepackage{mathtools}
\usepackage{subfig}
\usepackage{siunitx}
\usepackage[ruled,linesnumbered]{algorithm2e}
\usepackage[font=small]{caption}
\usepackage{booktabs}
\usepackage{makecell}
\usepackage{pgfplots}
\usepackage{pgfplotstable}
\usepackage{fancyhdr}

\usepackage{multirow} 
\usepackage{multicol} 
\usepackage{booktabs} 

\usepackage{soul} 

\usepackage{epstopdf}
\makeatletter
\let\NAT@parse\undefined
\makeatother
\usepackage[numbers]{natbib}
\usepackage[capitalise]{cleveref}

\newcommand{\wspace}[1]{\mathcal{W}_{\mathrm{#1}}}
\newcommand{\rspace}[1]{\mathbb{R}^{#1}}

\newcommand{\dims}{q}

\newcommand{\iden}{\mathbf{I}}

\newcommand{\ellipsoid}[4]{\mathcal{E}^{#1}_{#2}(#3, #4)}
\newcommand{\gaussian}{\mathcal{N}}
\newcommand{\mean}{\boldsymbol{\mu}}
\newcommand{\cov}{\boldsymbol{\Sigma}}
\newcommand{\pdens}{f}

\newcommand{\rob}[1]{\mathcal{P}_{#1}}
\newcommand{\rpos}{\mathbf{p}}
\newcommand{\rshape}{\mathbf{P}}
\newcommand{\rvar}{P}
\newcommand{\rrot}{\mathbf{R}}
\newcommand{\rscale}{\mathbf{S}}

\newcommand{\pcld}{\mathcal{Z}}
\newcommand{\spos}{\mathbf{z}}
\newcommand{\svar}{Z}

\ifx\BlackBox\undefined
\newcommand{\BlackBox}{\rule{1.5ex}{1.5ex}}  
\fi
\ifx\QED\undefined
\def\QED{~\rule[-1pt]{5pt}{5pt}\par\medskip}
\fi
\ifx\proof\undefined
\newenvironment{proof}{\par\noindent{\em Proof:\ }}{\hfill\BlackBox\\[.0mm]}
\fi
\ifx\theorem\undefined
\newtheorem{theorem}{Theorem}
\fi
\ifx\example\undefined
\theoremstyle{definition}
\newtheorem{example}{Example}
\fi
\ifx\property\undefined
\theoremstyle{definition}
\newtheorem{property}{Property}
\fi
\ifx\lemma\undefined
\newtheorem{lemma}{Lemma}
\fi
\ifx\proposition\undefined
\newtheorem{proposition}{Proposition}
\fi
\ifx\remark\undefined
\theoremstyle{remark}
\newtheorem{remark}{Remark}
\fi
\ifx\corollary\undefined
\newtheorem{corollary}{Corollary}
\fi
\ifx\definition\undefined
\newtheorem{definition}{Definition}
\fi
\ifx\conjecture\undefined
\newtheorem{conjecture}{Conjecture}
\fi
\ifx\axiom\undefined
\newtheorem{axiom}[theorem]{Axiom}
\fi
\ifx\claim\undefined
\newtheorem{claim}[theorem]{Claim}
\fi
\ifx\assumption\undefined
\newtheorem{assumption}{Assumption}
\fi

\usepackage[utf8]{inputenc}
\usepackage{pgfplots}
\usepgfplotslibrary{groupplots,dateplot}
\usetikzlibrary{patterns,shapes.arrows,arrows.meta}
\pgfplotsset{compat=newest,scaled x ticks=false}

\usepackage{tikz}
\usetikzlibrary{external,positioning}
\usepgfplotslibrary{external}
\ifthenelse{\equal{\arxivmode}{false}}
{

}
{

}

\definecolor{green}{RGB}{0,128,0}

\title{Distance and Collision Probability Estimation from Gaussian Surface Models}

\author{Kshitij Goel and Wennie Tabib
\thanks{The authors are with The Robotics Institute, Carnegie Mellon University, Pittsburgh, PA 15213 USA
(email: \{\texttt{kshitij,wtabib}\}\texttt{@cmu.edu}).%
}
}

\begin{document}
\maketitle
\thispagestyle{empty}
\pagestyle{empty}

\begin{abstract}
This paper describes methodologies to estimate the collision probability,
Euclidean distance and gradient between a robot and a surface, without
explicitly constructing a free space representation. The robot is assumed to be
an ellipsoid, which provides a tighter approximation for navigation in cluttered
and narrow spaces compared to the commonly used spherical model. Instead of
computing distances over point clouds or high-resolution occupancy grids, which
is expensive, the environment is modeled using compact Gaussian mixture models
and approximated via a set of ellipsoids. A parallelizable strategy to
accelerate an existing ellipsoid-ellipsoid distance computation method is
presented. Evaluation in 3D environments demonstrates improved performance over
state-of-the-art methods. Execution times for the approach are within a few
microseconds per ellipsoid pair using a single-thread on low-power embedded
computers.
\end{abstract}

\section{Introduction}\label{sec:intro}
\begin{figure}
  \ifthenelse{\equal{\arxivmode}{false}}%
  {
  \subfloat[Point Cloud\label{sfig:pcld_3d}]{\includegraphics[width=0.5\columnwidth,trim=480 200 640 75, clip]{figures/pcld_3d.eps}}%
  \subfloat[3D Ellipsoids\label{sfig:gmm}]{\includegraphics[width=0.5\columnwidth,trim=480 200 620 75, clip]{figures/gmm.eps}}\\
  \subfloat[Euclidean Distance Field\label{sfig:prop_df_3d}]{\includegraphics[width=0.5\columnwidth,trim=480 200 250 75, clip]{figures/prop_df_3d.eps}}%
  \subfloat[Collision Probability\label{sfig:prop_bl_3d}]{\includegraphics[width=0.5\columnwidth,trim=480 200 250 75, clip]{figures/prob_bl_3d.eps}}
  }%
  {
  \subfloat[Point Cloud\label{sfig:pcld_3d}]{\includegraphics[width=0.5\columnwidth,trim=480 200 640 75, clip]{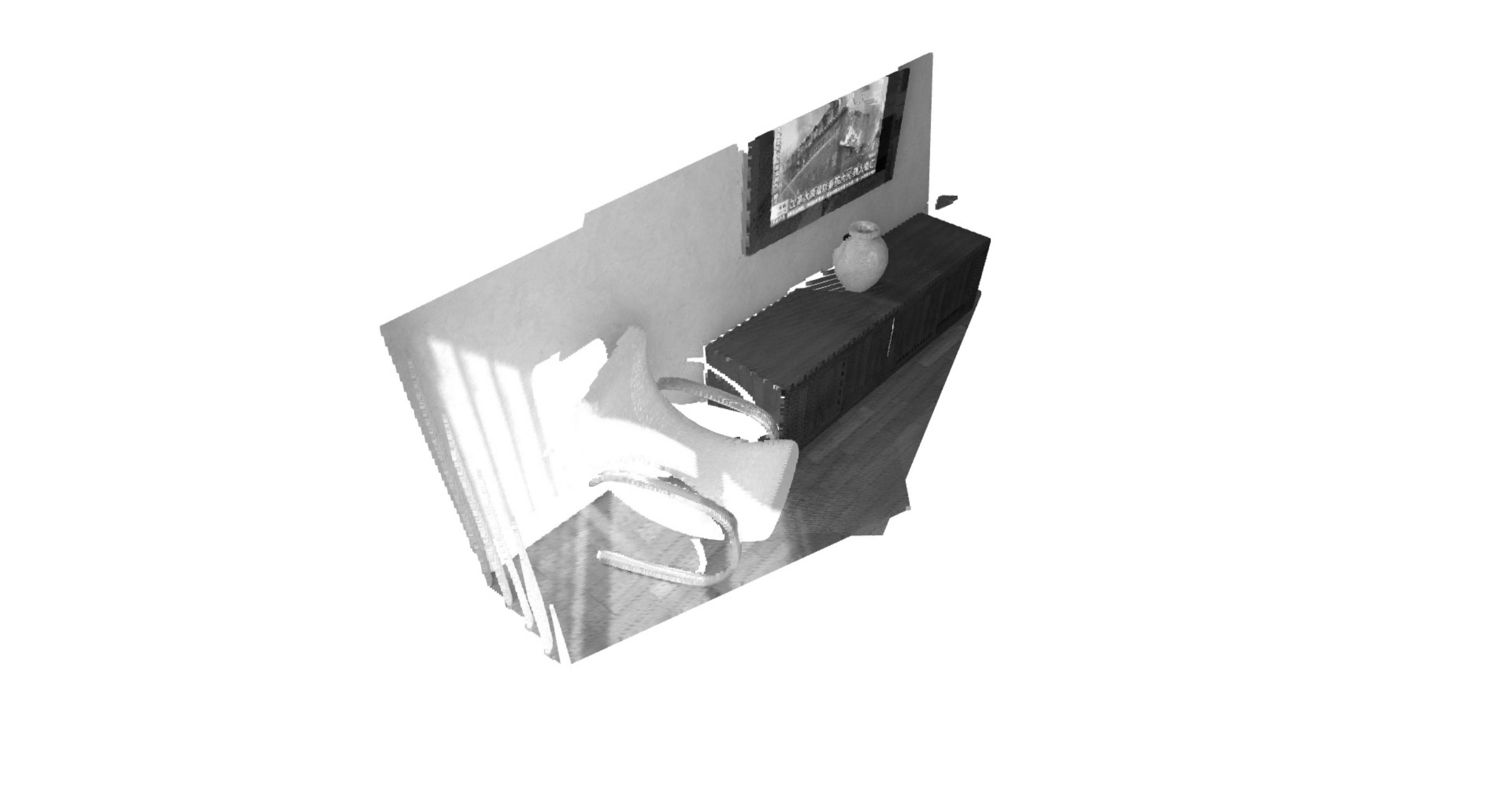}}%
  \subfloat[3D Ellipsoids\label{sfig:gmm}]{\includegraphics[width=0.5\columnwidth,trim=480 200 620 75, clip]{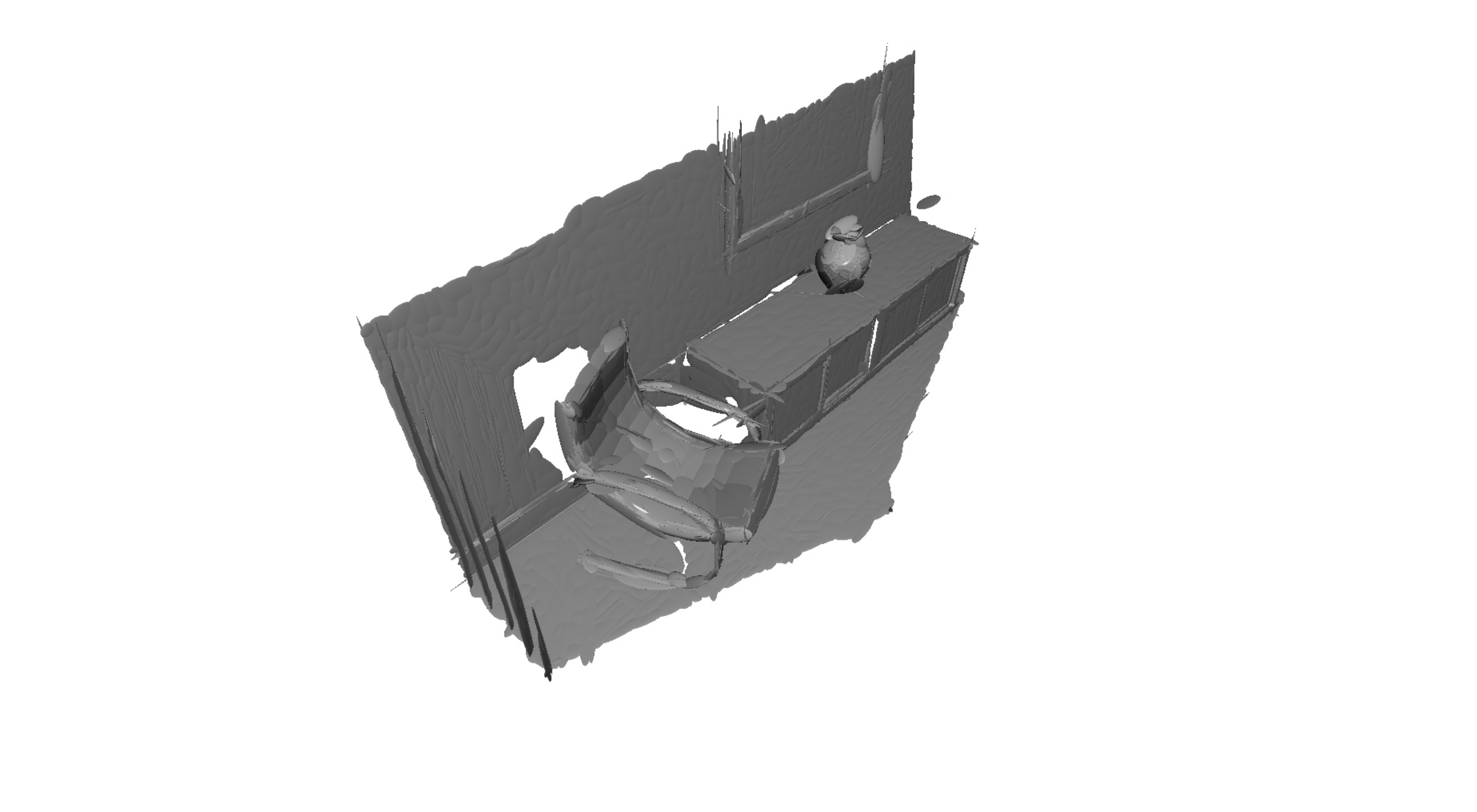}}\\
  \subfloat[Euclidean Distance Field\label{sfig:prop_df_3d}]{\includegraphics[width=0.5\columnwidth,trim=480 200 250 75, clip]{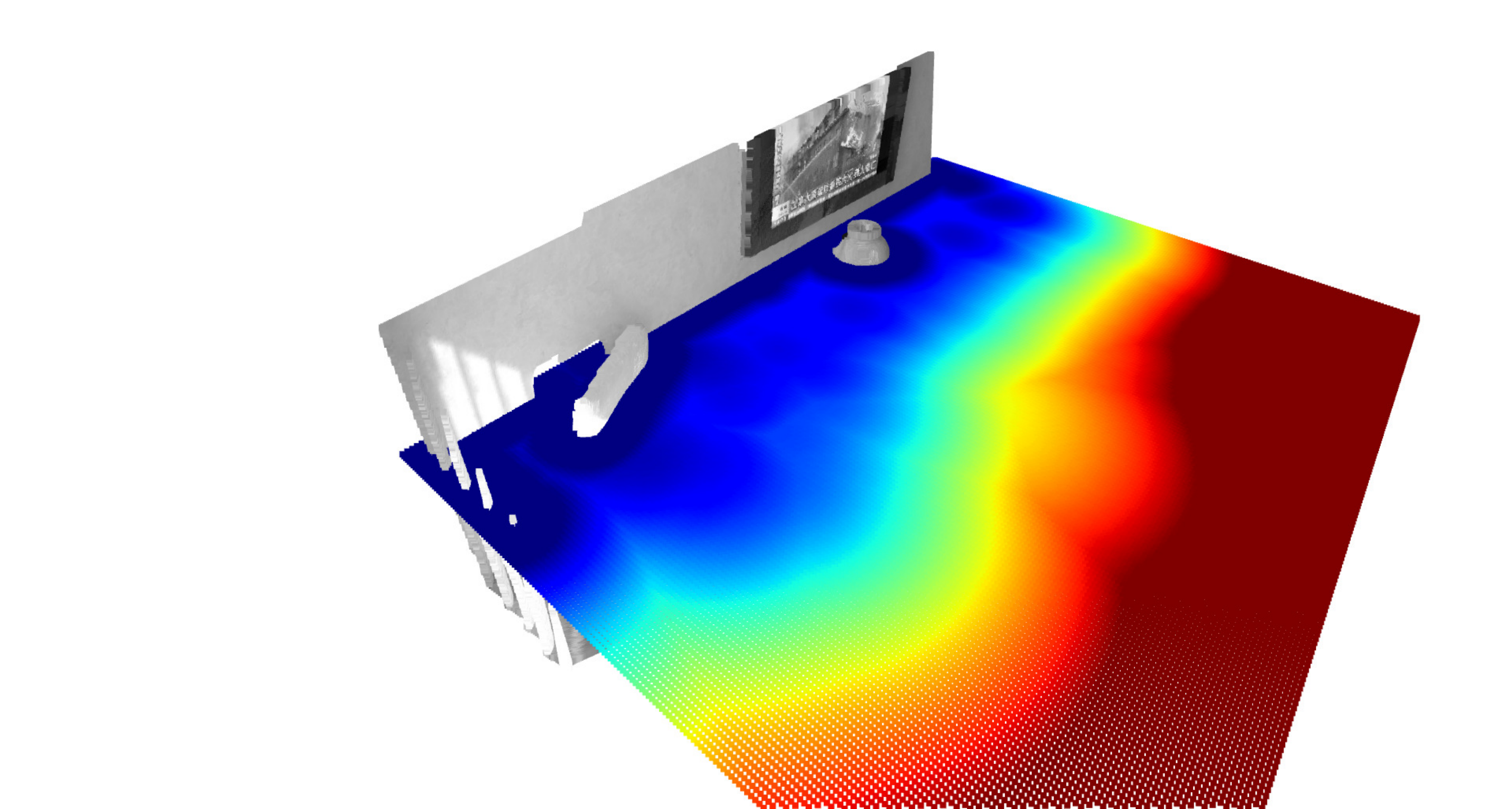}}%
  \subfloat[Collision Probability\label{sfig:prop_bl_3d}]{\includegraphics[width=0.5\columnwidth,trim=480 200 250 75, clip]{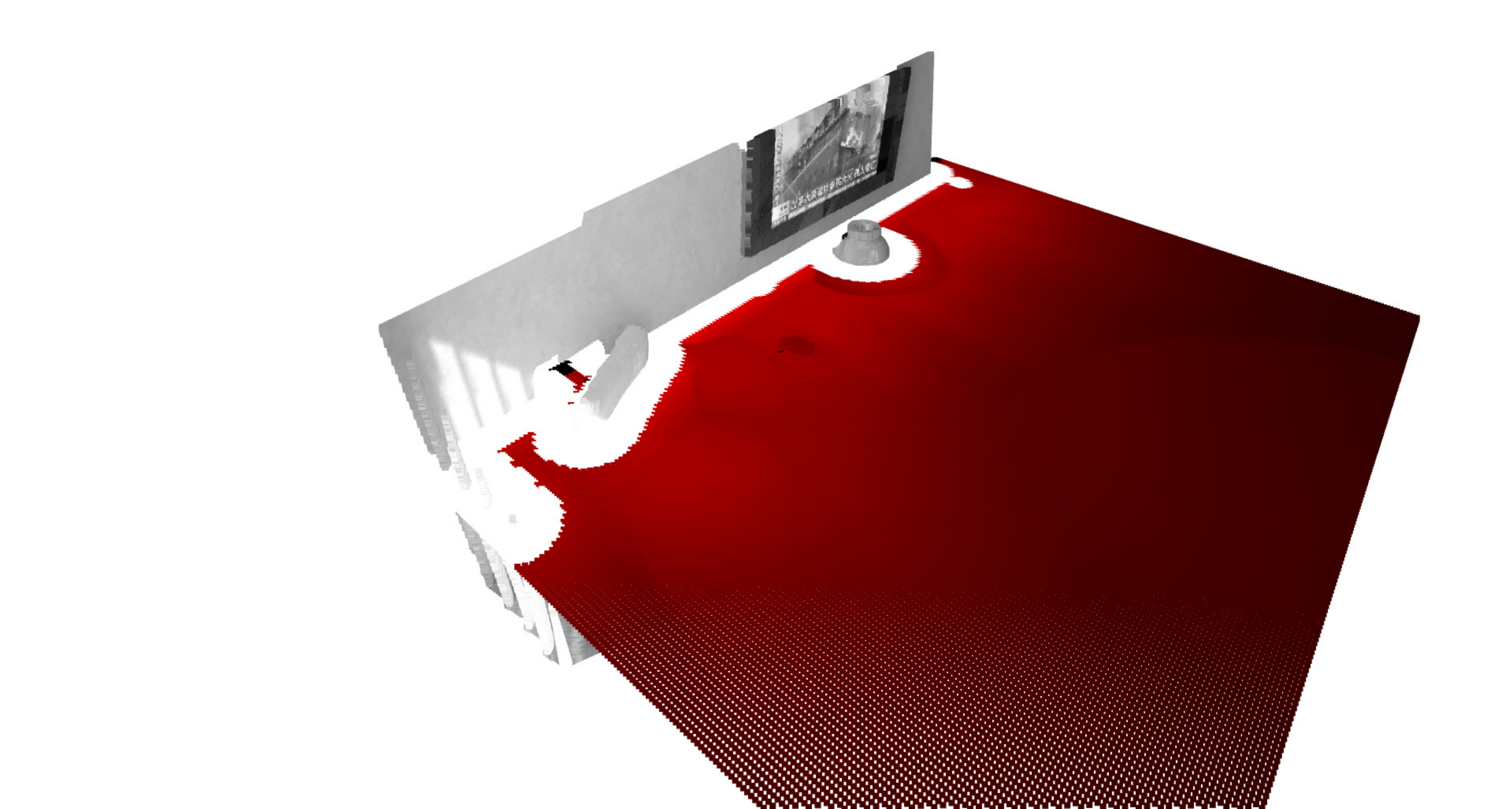}}
  }
  \caption{\label{glory-shot}This work contributes methods to estimate continuous-space
  collision probability, Euclidean distance and gradient of an ellipsoidal
  robot body model from a Gaussian surface model (GSM). The 3D point cloud shown
  in~\protect\subref{sfig:pcld_3d} is approximated with a GSM,
  shown as a set of ellipsoids in~\protect\subref{sfig:gmm}.~\protect\subref{sfig:prop_df_3d}
  The Euclidean distance over a 2D slice predicted by
  the proposed approach is shown as a heatmap (increasing
  distances from \textcolor{blue}{blue} to \textcolor{red}{red}).~\protect\subref{sfig:prop_bl_3d}
  The collision probability values (decreasing from \textcolor{red}{red} to
  black, $1.0$ in white regions) over the same 2D slice when the robot position
  is uncertain. 
  }
  \vspace{-0.5cm}
\end{figure}

Euclidean distance fields and their gradients (often computed
using finite differencing) are used for collision
avoidance~\citep{oleynikova_voxblox_2017} and optimization-based
motion planning~\citep{ratliff_chomp_2009}, respectively.
These values are calculated by fusing point clouds
acquired by range sensors (e.g., LiDARs and depth cameras) and
processed on-board the robot.
To enable safe navigation outside the sensor field-of-view (FoV), the robot
must maintain a map comprised of past fused
observations.
This is especially important when limited FoV sensors,
like depth cameras, are used onboard the
robot~\citep{tordesillas_faster_2022}. Existing Euclidean distance and
gradient estimation methods create a spatially discretized map and compute distance
and gradient values in each
cell via a Breadth-First Search~\citep{han_fiesta_2019}, which is computationally
expensive in large workspaces or when small grid cells are employed~\citep{chen_gpu-accelerated_2022}.
To overcome the limitations of cell based methods, Gaussian
Process (GP) based methods were developed, which implicitly represent the
surface and enable continuous-space Euclidean distance and gradient
calculation~\citep{lee_online_2019,wu_log-gpis-mop_2023,le_gentil_accurate_2024}.
A GP is regressed from raw point cloud data and used to infer
the Euclidean distance and gradient at
any test point in the workspace.
A challenge with these methods is that the training time scales
as the size of the input data increases, so optimization-based
subsampling must be introduced to enable spatial
scalability~\citep{wu_pseudo_2023}.
In contrast, Gaussian surface models (GSMs) (e.g.,
mixture models~\citep{eckart_accelerated_2016,srivastava_efficient_2019,omeadhra_variable_2019,li_memory-efficient_2022}
and 3D splatting works~\citep{dhawale_efficient_2020,keselman_approximate_2022,kerbl_3d_2023})
provide relatively compressed and high-fidelity point cloud models.
This work leverages the geometric interpretation of
GSMs to calculate continuous-space Euclidean distance and gradients (\cref{glory-shot}\footnote{A video is available at \url{https://youtu.be/lRBoQNxd65g}.}).

Collision probabilities are used for motion planning under robot position
uncertainty~\citep{du_toit_probabilistic_2011,bry_rapidly-exploring_2011}.
Like Euclidean distance and gradient field methods, existing
methods that estimate collision probability from a surface
point cloud utilize probabilistic occupancy queries from a
discrete~\citep{heiden_planning_2017,laconte_lambda-field_2019} or
continuous surface representation~\citep{francis_functional_2020}.
These maps require storing free space
cells~\citep{elfes_using_1989,hornung_octomap_2013,agha-mohammadi_confidence-rich_2019}
or raw point clouds~\citep{ramos_hilbert_2016}, leading to high
memory usage. GSMs can alleviate these challenges by effectively compressing point clouds into
a finite set of ellipsoids. However,
existing occupancy estimation frameworks leveraging GSMs either fit
Gaussian components in free
space~\citep{srivastava_efficient_2019,li_gmmap_2024} or Monte
Carlo ray trace through a local discrete
grid~\citep{omeadhra_variable_2019,tabib_real-time_2019}. Robot
pose uncertainty is also not considered. This
paper bridges these gaps by proposing a collision probability
estimation method, which uses only the surface Gaussians and
accounts for Gaussian uncertainty in the robot position.

All the aforementioned techniques require approximating the shape of the robot.
Spherical approximations are widely used because inflating the obstacles
with the robot size for the purposes of collision avoidance results in
a simple configuration space (C-space)~\citep{latombe_robot_1991}. However,
recent results in ellipsoidal approximations demonstrate superior planning
performance, especially in cluttered and narrow
environments~\citep{ruan_efficient_2023} for both
rigid~\citep{liu_search-based_2018} and articulated robots~\citep{ji_e-rrt_2023}.
These approximations are also used for human bodies in a
dynamic environment~\citep{zhu_chance-constrained_2019,liu_tight_2023}.
Consequently, in this work the proposed methods use an ellipsoidal robot body approximation.

\textit{Contributions}: Given an ellipsoidal robot body model and a set of Gaussians representing
the surface point cloud, this paper contributes (\cref{glory-shot}) methods to compute the continuous-space
\begin{enumerate}
  \item Euclidean distance estimate between the robot body and surface;
  \item approximate gradient of this distance; and
  \item upper bound on collision probability when the robot position is a Gaussian random variable.
\end{enumerate}
These methods are evaluated using 3D surface point clouds from simulation and
real-world environments. The software associated with this paper is open-sourced
for the benefit of the research community\footnote{Release \texttt{0.0.1} of
\href{https://github.com/gira3d/gira3d-map-ops}{\texttt{gira3d-map-ops}}.}.

\Cref{sec:approach} details the proposed methods. The evaluation is presented
in~\cref{sec:results}. Concluding remarks and directions for future work are
in~\cref{sec:conclusion}.


\section{Methodology}\label{sec:approach}
Starting with the preliminary information, the problem
statement is provided in~\cref{ssec:ps}. The proposed methods are detailed
in~\cref{ssec:cedf} and~\cref{ssec:cpupu}. In this section, small letters are
scalars (e.g. $x$, $y$), bolded small letters are vectors (e.g. $\mathbf{x}$,
$\mathbf{y}$), capital letters are random variables (e.g. $X$, $Y$), capital
bolded letters are matrices (e.g. $\mathbf{X}$, $\mathbf{Y}$), and calligraphic
letters are sets (e.g. $\mathcal{X}, \mathcal{Y}$).

A solid ellipsoid of dimension $\dims$, center $\rpos
\in \rspace{\dims}$, and shape $\rshape \in \rspace{\dims
\times \dims}$ is the quadratic form inequality
\begin{equation*}
  \ellipsoid{}{}{\rpos}{\rshape} = \{ \mathbf{q} \in \rspace{\dims} \mid
  (\mathbf{q} - \rpos)^{\top} \rshape (\mathbf{q} - \rpos) \leq 1 \}.
\end{equation*}
In this work $\dims = 3$, but the proposed methods are amenable to $q = 2$.
The eigen decomposition of  $\rshape$ enables determining the rotation and scale
of the ellipsoid. If the eigen decomposition of $\rshape$ is given by $\rshape =
\rrot \rscale \rrot^{\top}$, then $\rrot$ is an orthogonal matrix that provides
the ellipsoid's principal axes rotation and $\rscale$ is a diagonal matrix where
entries are the inverse squares of the semi-principal axes lengths. The
positive definite matrix $\rshape^{-1}$ is called the shape
matrix of the ellipsoid~\citep{halder_parameterized_2018}.

The region inside and on a probability isocontour of the probability density function
for a $\dims$-variate Gaussian random variable $X$ with mean vector
$\mean_X$ and covariance matrix $\cov_X$ can be geometrically interpreted as
an ellipsoid~\citep[Ch. 2]{bishop_pattern_2006}. The $l$-th level isocontour is
given by the solid ellipsoid $\ellipsoid{}{X}{\mean_X}{\rrot \boldsymbol{\Gamma}
\rrot^{\top}}$. The rotation matrix $\rrot$ contains the eigenvectors of
$\cov_{X}^{-1}$ as columns.  The entries of the diagonal matrix
$\boldsymbol{\Gamma}$ are given by $1 / l^2$ times the eigenvalues of
$\cov_{X}^{-1}$. For reference, $l = 3$ and $4$ provide $99.7\%$ and $99.95\%$
coverage bounds, respectively, on the points modeled by the random variable $X$.

If the probability density function is instead given by a Gaussian Mixture Model
(GMM), then the probability isocontours can be approximated as a set of
ellipsoids with one ellipsoid per component of the mixture.

\subsection{Problem Statement\label{ssec:ps}}
A robot $\rob{}$ is equipped with a range sensor (e.g. depth camera, LiDAR,
etc.) and it can be moved in a workspace $\wspace{} \subseteq \rspace{\dims}$.
The
region inside the workspace occupied by the robot is modeled using the solid
ellipsoid $\ellipsoid{}{\rob{}}{\rpos}{\rshape}$.  The parameters $\rpos$ and
$\rshape$ can be determined using the Lowner-John ellipsoid
fit~\citep{rimon_obstacle_1997}. The center $\rpos$ can be uncertain. This
uncertainty is modeled using a continuous multivariate Gaussian variable $\rvar$
with the probability density function $\gaussian(\rpos; \mean_{\rvar},
\cov_{\rvar})$. In practice, these uncertainty estimates may be provided by an
external state estimation system~\citep{mourikis_multi-state_2007}.

The onboard range sensor provides a stream of point cloud measurements
$\pcld \subseteq \wspace{s}$ of the surfaces $\wspace{s}$ in the workspace. Each
point $\mathbf{z} \in \rspace{\dims}$ in $\pcld$ is assumed to be an independent and identically distributed
sample of a random variable $\svar$ with the probability density function given
by the GMM
\begin{equation}
  \pdens_{\svar} = \sum_{m=1}^{M} \pi_m \gaussian(\spos; \mean^{m}_{\svar}, \cov^{m}_{\svar})
  \label{eq:gmm}
\end{equation}
where $\pi_m$, $\mean^{m}_{\svar}$, and $\cov^{m}_{\svar}$ are the weight, mean, and covariance
of the $m$-th Gaussian component in an $M$-component GMM. This GMM can be
obtained from many recent point cloud modeling methods that utilize stochastic
gradient-descent~\citep{keselman_approximate_2022,kerbl_3d_2023} (i.e., Gaussian Splatting), geometric
region growing~\citep{dhawale_efficient_2020}, Expectation
Maximization~\citep{goel_incremental_2023,srivastava_efficient_2019}, and scan
line segmentation~\citep{li_memory-efficient_2022}.  The proposed methods in
this work only depend on the geometric interpretation of the GMMs and thus do
not require the GMMs to be created using a specific method.

Given this information, the following problems are addressed in this work:
\begin{enumerate}
  \item Estimate the closest distance between
  $\ellipsoid{}{\rob{}}{\rpos}{\rshape}$ and $\pdens_{\svar}$.  Estimate
  the gradient of this distance at any location. The solution must not
  depend on a discretization of the space (\cref{ssec:cedf}).
  \item Estimate the collision probability between
  $\ellipsoid{}{\rob{}}{\rpos}{\rshape}$ and $\pdens_{\svar}$ when the ellipsoid
  center position is uncertain
  (\cref{ssec:cpupu}).
\end{enumerate}

\subsection{Continuous Euclidean Distance Field\label{ssec:cedf}}
The $m$-th component of the GMM density $\pdens_{\svar}$ from~\cref{eq:gmm} can be interpreted as an
ellipsoid. If it were possible to efficiently calculate the
distance of the robot to all $M$ components, we could compute the minimum distance
of the robot to the surface GMM. Thus, we require a method to find the closest
distance between two ellipsoids. Unfortunately, this distance cannot be derived
as a closed-form expression~\citep{lin_distance_2002}.

Many optimization-based approaches have been proposed for this
purpose~\citep{lin_distance_2002,rimon_obstacle_1997,tracy_differentiable_2023}.
In this work, the method by~\citet{rimon_obstacle_1997} is leveraged because it
formulates the optimization as an eigenvalue problem. However, the
expressions stated in their work require several matrix inversions and square
roots. It is not obvious how to avoid these calculations and leverage numerically
stable and faster alternatives (e.g. linear system solvers, decompositions) without
introducing approximations. The
following proposition restates the result from~\citep{rimon_obstacle_1997} but
removes the need to explicitly calculate matrix inversions and square roots.

\begin{proposition}[Deterministic Ellipsoid Distance~\citep{rimon_obstacle_1997}\footnote{The typographical errors in
the expressions for $\tilde{\mathbf{b}}$ and $\mathbf{y}^*$ in~\citep[Prop.~3.2]{rimon_obstacle_1997}
have been corrected here.}]
\label{lem:dist}
Consider two ellipsoids $\ellipsoid{}{1}{\mathbf{b}}{\mathbf{B}}$ and
$\ellipsoid{}{2}{\mathbf{c}}{\mathbf{C}}$. The centers and shapes of these
ellipsoids are perfectly known. Let $\mathbf{B} = \hat{\mathbf{B}}
\boldsymbol{\Lambda}_B \hat{\mathbf{B}}^{\top}$ and $\mathbf{C} =
\hat{\mathbf{C}} \boldsymbol{\Lambda}_C \hat{\mathbf{C}}^{\top}$ be the eigen
decompositions of $\mathbf{B}$ and $\mathbf{C}$ respectively. Consequently,
$\mathbf{B}^{1/2} = \hat{\mathbf{B}}
\boldsymbol{\Lambda}^{1/2}_B \hat{\mathbf{B}}^{\top}$, $\mathbf{B}^{-1} = \hat{\mathbf{B}}
\boldsymbol{\Lambda}^{-1}_B \hat{\mathbf{B}}^{\top}$, $\mathbf{B}^{-1/2} = \hat{\mathbf{B}}
\boldsymbol{\Lambda}^{-1/2}_B \hat{\mathbf{B}}^{\top}$, and $\mathbf{C}^{-1} = \hat{\mathbf{C}}
\boldsymbol{\Lambda}^{-1}_C \hat{\mathbf{C}}^{\top}$.
Let $\lambda$ be the minimal eigenvalue\footnote{The minimal eigenvalue is equal to the real part of the eigenvalue with
the lowest real part amongst all eigenvalues (real or complex). This notion is required because
the eigenvalues of $\mathbf{M}_1$ and $\mathbf{M}_2$ may be complex numbers.} of the $2q \times 2q$ matrix
\begin{equation*}
  \mathbf{M}_1 = \begin{bmatrix} \tilde{\mathbf{C}} & - \iden_q \\ -\tilde{\mathbf{c}}\tilde{\mathbf{c}}^{\top} & \tilde{\mathbf{C}} \end{bmatrix}
\end{equation*}
such that $\iden_q$ is an identity matrix of order $q$,
$\tilde{\mathbf{C}} = \mathbf{B}^{1/2} \mathbf{C}^{-1} \mathbf{B}^{1/2}$, and $\tilde{\mathbf{c}}$
is the solution of the linear system
\begin{equation}
  (\mathbf{B}^{-1/2} \hat{\mathbf{Q}} \boldsymbol{\Lambda}^{1/2}_Q \hat{\mathbf{Q}}^{\top}) \tilde{\mathbf{c}} = \mathbf{c} - \mathbf{b}
\end{equation}
where $\hat{\mathbf{Q}} \boldsymbol{\Lambda}_Q \hat{\mathbf{Q}}^{\top}$
is the eigen decomposition of the real symmetric matrix $\mathbf{B}^{-1/2} \mathbf{C} \mathbf{B}^{-1/2}$.
Let $\mu$ be the minimal eigenvalue of the $2q \times 2q$ matrix
\begin{equation*}
  \mathbf{M}_2 = \begin{bmatrix} \mathbf{B}^{-1} & - \iden_q \\ -\tilde{\mathbf{b}}\tilde{\mathbf{b}}^{\top} & \mathbf{B}^{-1} \end{bmatrix}
\end{equation*}
such that $\tilde{\mathbf{b}} = -\lambda\mathbf{B}^{-1/2}\boldsymbol{\alpha}$ and
$\boldsymbol{\alpha}$ is the solution to the linear system
\begin{equation}
  \{\mathbf{B}^{-1/2}(\lambda \iden_q - \tilde{\mathbf{C}}) \mathbf{B}^{1/2}\} \boldsymbol{\alpha} = \mathbf{c} - \mathbf{b}.
\end{equation}
Given these quantities, the closest distance estimate $d(\mathcal{E}_1, \mathcal{E}_2)$ between
$\mathcal{E}_1$ and $\mathcal{E}_2$ is
\begin{equation}
  d(\mathcal{E}_1, \mathcal{E}_2) = \| \mathbf{d}^{*} \|,
  \label{eq:ell-dist}
\end{equation}
where $\|.\|$ denotes the L2-norm of a vector and $\mathbf{d}^*$ is the solution to the linear system
\begin{equation}
  (\mu \iden_q - \mathbf{B}^{-1}) \mathbf{d}^{*} = - \mu \lambda \boldsymbol{\alpha}.
\end{equation}
\end{proposition}

Calculating matrix inverses and square roots of diagonal matrices like
$\boldsymbol{\Lambda}_B$ only requires inverse and square root operations on
their scalar diagonal entries as opposed to full matrix operations. Therefore, there are
no explicit matrix inversion or matrix square root calculations required in~\cref{lem:dist}.

Using~\cref{lem:dist}, the distance between the robot body ellipsoid
$\ellipsoid{}{\rob{}}{\rpos}{\rshape}$ and the surface model $\pdens_{\svar}$ can be computed
by computing the minimum over all $M$ components of $\pdens_{\svar}$:
\begin{equation}
  d^*(\ellipsoid{}{\rob{}}{\rpos}{\rshape}, \pdens_{\svar}) = \min_{m} d(\ellipsoid{}{\rob{}}{\rpos}{\rshape}, \mathcal{E}_m)
  \label{eq:gmm-dist}
\end{equation}
where $d(.)$ is the distance function from~\cref{eq:ell-dist} and $\mathcal{E}_m$ denotes the ellipsoid
corresponding to the $m$-th Gaussian component in the GMM density $\pdens_{\svar}$. This ellipsoid
can be constructed for isocontours of the Gaussian component.

To calculate the gradient,~\citet{rimon_obstacle_1997} suggest deriving the
analytical gradient of~\cref{eq:ell-dist} as it is differentiable. However,
computing this gradient requires several matrix multiplications and inversions.
To save computational resources onboard robots, an approximation is leveraged.
From~\cref{eq:gmm-dist}, we also get the ellipsoid on the surface,
$\mathcal{E}^{*}_{m}$, that is closest to the robot.
The gradient vector is approximated using the position vector $\mathbf{d}^*$
(from~\cref{eq:ell-dist}) for $\mathcal{E}^{*}_{m}$ and $\mathcal{E}_{\rob{}}$,
\begin{equation}
  \nabla d^{*}(\ellipsoid{}{\rob{}}{\rpos}{\rshape}, \pdens_{\svar}) = \frac{\mathbf{d}^{*}}{\sqrt{\mathbf{d}^{{*}^{\top}}\mathbf{d}^{*}}}.
  \label{eq:gmm-grad}
\end{equation}

Under the limiting condition where the number of components $M$ is equal to the
number of points in the point cloud $\pcld$ such that each point in the point
cloud is represented with a Gaussian component, these methods are
equivalent to a nearest-neighbor search and the distance (\cref{eq:gmm-dist})
and gradient (\cref{eq:gmm-grad}) formulations are exact up to point
cloud resolution. As $M$ decreases, the error in $d^{*}$ and $\nabla d^{*}$
increases relative to the ground truth value.

The computation in~\cref{eq:gmm-dist} scales linearly with the number of
components $M$ in the GMM. Local submap extraction approaches such as hash
maps~\citep{goel_incremental_2023} or spatial partitioning data structures such
as VDB~\citep{museth_vdb_2013} may enable improved scalability as $M$
increases.

\begin{figure}
  \centering
  \includegraphics[width=\columnwidth,trim=0 500 1350 0,clip]{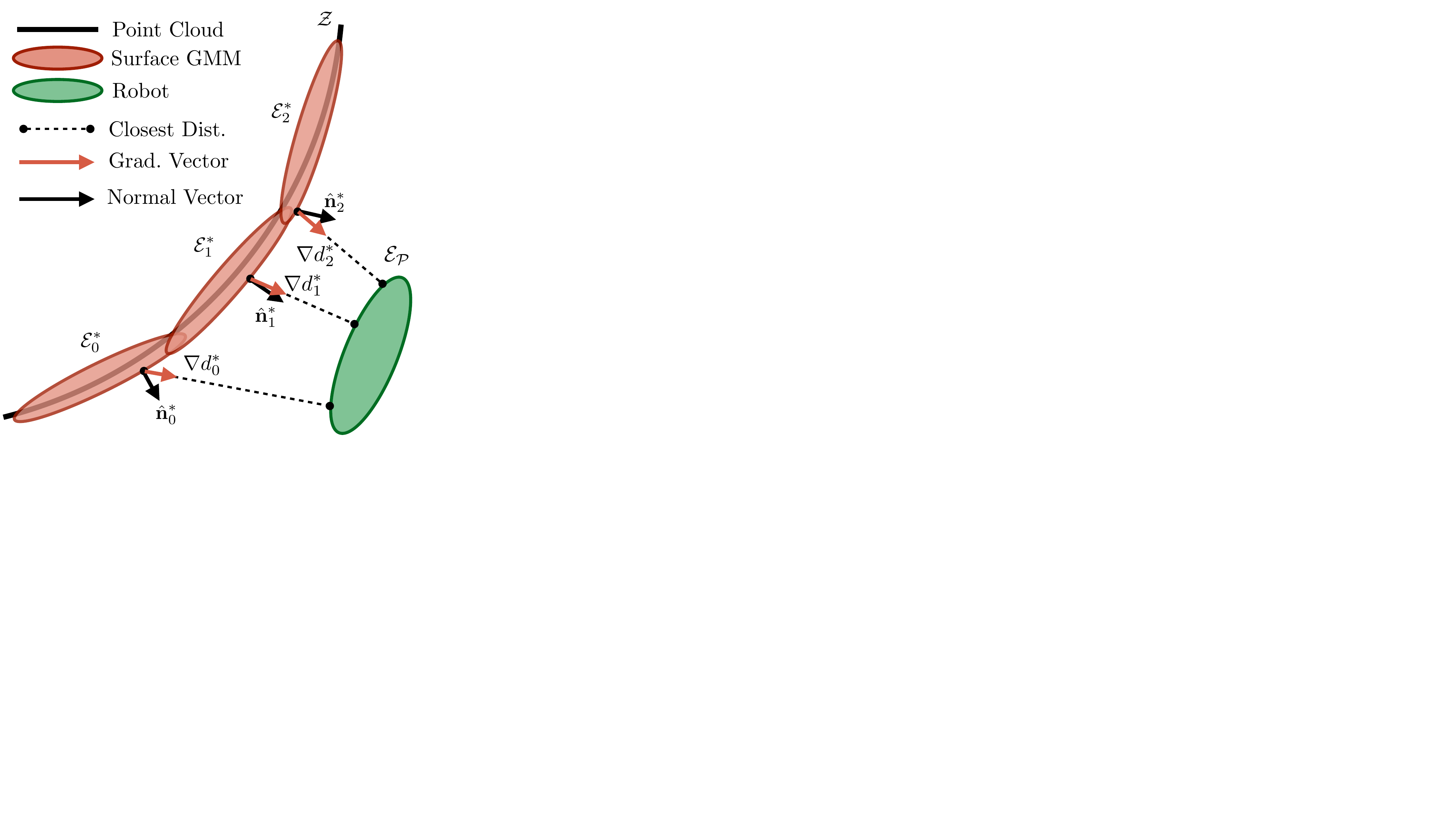}
  \caption{\label{illus:blending}Illustration of the quantities required for blending weights calculation in collision
  probability estimation (\cref{ssec:cpupu}). For each of the ellipsoids on the surface, the blending
  weight $w_k$ is the dot product of the distance gradient $\nabla d^{*}_k$ and the normal $\hat{\mathbf{n}}^{*}_k$.
  }
\end{figure}

\subsection{Collision Probability Under Position Uncertainty\label{ssec:cpupu}}
The following corollary to~\cref{lem:dist} will be used in this section.
\begin{corollary}[Deterministic Ellipsoid Collision Check~\citep{thomas_exact_2022}]
\label{cor:coll-check}
If it is desired to only check whether $\mathcal{E}_1$ and $\mathcal{E}_2$
collide (under deterministic conditions), an explicit calculation of~\cref{eq:ell-dist} is not required. When
$\mathcal{E}_1$ and $\mathcal{E}_2$ touch or
intersect,~\citet{thomas_exact_2022} show that
$\mathbf{y}^{\top}\mathbf{D}^{\top}\mathbf{B}\mathbf{D}\mathbf{y} \leq 1 /
\lambda^{2}$, where $\mathbf{y} = \mathbf{c} - \mathbf{b}$, $\mathbf{D} =
\mathbf{B}^{-1/2}(\lambda \iden_q - \tilde{\mathbf{C}})^{-1} \mathbf{B}^{1/2}$,
and other quantities are as calculated in~\cref{lem:dist}. We simplify this
inequality to
\begin{equation}
 \mathbf{y}^{\top} \mathbf{B}^{1/2} \mathbf{A}^{-1} \mathbf{B}^{1/2} \mathbf{y} \leq 1 / \lambda^{2}
 \label{eq:coll-check-simpl-1}
\end{equation}
where $\mathbf{A} = (\lambda \iden_q - \tilde{\mathbf{C}})^{2}$ is a
real symmetric matrix. It is shown in~\citep[Thm.~2]{rimon_obstacle_1997} that
when the center of ellipsoid $\mathcal{E}_1$, $\mathbf{b}$, lies outside of
$\mathcal{E}_2$\footnote{This assumption does not lead to a loss of generality because
when $\mathbf{b}$ lies inside $\mathcal{E}_2$, the ellipsoids are definitely intersecting.},
$\lambda$ is always negative. Since $\tilde{\mathbf{C}} = \mathbf{B}^{1/2} \mathbf{C}^{-1} \mathbf{B}^{1/2}$
with $\mathbf{B} \succ 0$ and $\mathbf{C} \succ 0$, it follows that $\tilde{\mathbf{C}} \succ 0$. Therefore, the matrix $\lambda \iden_q -
\tilde{\mathbf{C}} \prec 0$, which implies $\mathbf{A} \succ 0$ (i.e., $\mathbf{A}$ is
positive definite). Therefore, the Cholesky decomposition of $\mathbf{A}$ can be used to efficiently
compute~\cref{eq:coll-check-simpl-1} without explicit matrix inversions or square roots.
\end{corollary}

Let $\mathbf{v} = \mathbf{y}^{\top} \bar{\mathbf{A}} \mathbf{y}$ where
$\bar{\mathbf{A}} = \mathbf{B}^{1/2} \mathbf{A}^{-1} \mathbf{B}^{1/2}$.
When the center $\mathbf{b}$ of the ellipsoid $\mathcal{E}_1$ is a Gaussian distributed random
variable with density $\gaussian(\mathbf{b}; \mean_B, \cov_B)$, $\mathbf{v}$
becomes a random variable as both $\mathbf{y}$ and $\bar{\mathbf{A}}$ depend on $\mathbf{b}$.
Under a conservative assumption that $\bar{\mathbf{A}}$ and $\lambda$ are
deterministic (calculated using the mean $\mean_B$)\footnote{\citet{liu_tight_2023} propose
leveraging the Minkowski sum of ellipsoids to relax this assumption. However,
it is computationally difficult to obtain a tight approximation of this sum~\citep{halder_parameterized_2018}.},
it is proved in~\citep{thomas_exact_2022} that
$P_B(\mathcal{E}_1, \mathcal{E}_2) \equiv P_B(v \leq 1 / \lambda^2)$
and an upper bound on the probability can be estimated via
\begin{equation}
  P_B(v \leq 1 / \lambda^2) \leq \frac{\eta \sqrt{\mathbb{V}[v]}}{\mathbb{E}[v] + \eta \sqrt{\mathbb{V}[v]} - (1 / \lambda^2)}
  \label{eq:ell-prob}
\end{equation}
where $\mathbb{E}[v]$ and $\mathbb{V}[v]$ denote the expectation and variance of
$v$, $\lambda$ is as defined in~\cref{lem:dist}, and $\eta$ is a
constant. The
values $\mathbb{E}[v]$ and $\mathbb{V}[v]$ can be
exactly calculated using~\citep[Thm.~3.2b.2]{mathai_quadratic_1992}
\begin{align*}
&\mathbb{E}[v] = \text{tr}[\bar{\mathbf{A}} \cov_B] + (\mathbf{c} - \mean_B)^{\top} \bar{\mathbf{A}} (\mathbf{c} - \mean_B) \text{ and}\\
&\mathbb{V}[v] = 2 \text{tr}[(\bar{\mathbf{A}} \cov_B)^2] + 4 (\mathbf{c} - \mean_B)^{\top} \bar{\mathbf{A}} \cov_B \bar{\mathbf{A}} (\mathbf{c} - \mean_B)
\end{align*}
where $\text{tr}[.]$ denotes the trace of a matrix. Explicit matrix square roots are not required
in this upper bound calculation (\cref{lem:dist}). Estimation of the constant $\eta$ is difficult.
\citet{thomas_exact_2022} set this value to $0.25$ for all cases\footnote{There
is a typographical error in~\citep{thomas_exact_2022} which states that $\eta =
1$. The tight upper bound calculated in Section IV-D
of~\citep{thomas_exact_2022} is correct when $\eta = 0.25$.}. However, we found that in some cases
this value leads the denominator in~\cref{eq:ell-prob} to be negative (i.e.,
$\mathbb{E}[\mathbf{v}] + \eta \sqrt{\mathbb{V}[\mathbf{v}]} < (1 / \lambda^2)$). Therefore,
in practice when this denominator is negative, starting with $\eta = 0.25$, we keep increasing it by $0.5$
until the denominator turns positive. This is a valid approximation because $\eta$ is used to upper
bound the value of $\mathbf{v}$~\citep{thomas_exact_2022}; increasing $\eta$ makes this bound
loose for certain ellipsoid pairs but maintains the validity of~\cref{eq:ell-prob}.

To compute the probability in~\cref{eq:ell-prob} relative to the surface GMM $\pdens_{\svar}$, one
approach is to identify the surface ellipsoid closest to the robot
($\mathcal{E}^{*}_m$) from~\cref{eq:gmm-dist} and evaluate~\cref{eq:ell-prob}
between the robot ellipsoid $\ellipsoid{}{\rob{}}{\rpos}{\rshape}$ and
$\mathcal{E}^{*}_m$. However, this approach yields a non-smooth collision
probability field due to a rapid spatial change in $\mathcal{E}^{*}_m$. Such a
probability field may not be useful for optimization-based motion planning under
uncertainty~\citep{zhu_chance-constrained_2019}.

To mitigate this problem, we propose blending the collision probabilities of the
$K$ nearest surface ellipsoids from the robot ellipsoid. Such neighbors
can be efficiently queried from spatial partitioning data structures such as
KD-Tree~\citep{ichnowski_concurrent_2020}. The final probability is given by
\begin{equation}
  P^{*}(\ellipsoid{}{\rob{}}{\rpos}{\rshape}, \pdens_{\svar}) = \frac{1}{\sum_{k} w_k} \sum_{k} w_k P(\ellipsoid{}{\rob{}}{\rpos}{\rshape}, \mathcal{E}^{*}_k),
  \label{eq:gmm-prob}
\end{equation}
where $w_k$ denotes the blending weight of the $k$-th ellipsoid
$\mathcal{E}^{*}_k$ in the set of nearest $K$ ellipsoids $\{ \mathcal{E}^{*}_1,
\ldots, \mathcal{E}^{*}_K \}$. The blending weight $w_k$ is the dot product of
$\nabla d^{*}_k$ and the eigenvector directed towards the robot position,
$\hat{\mathbf{n}}^{*}_k$, corresponding to the minimum eigenvalue of the shape
matrix (\cref{illus:blending}). This dot product is low when the Gaussian component is
pointing away from the direction of distance gradient, thus lowering its contribution to the
final probability. The dot product will be negative in cases
when the angle between $\hat{\mathbf{n}}^{*}_k$ and $\nabla d^{*}_k$ is
in the range $[\pi/2, 3\pi / 2]$. For example, this situation can arise when ellipsoids
represent a sharp turn. In such cases, we ignore the contribution of the ellipsoid
to the collision probability (i.e., $w_k = 0$).


\section{Results \& Discussion}\label{sec:results}
This section presents an evaluation of the computational cost and accuracy of the proposed methods.

Targeting single-threaded operation on a CPU, the proposed methods are
implemented in C++ using the Eigen library~\citep{eigenweb} with the highest
level of optimization enabled in the GNU GCC compiler (\texttt{-march=native
-O3}). Python bindings are created using \texttt{pybind11} for easy integration,
testing, and visualization. The performance of this implementation is measured
on several desktop and embedded platform CPUs to reflect the applicability of
the proposed methods for a wide range of robots (\cref{ssec:perf-compute}).

The accuracy of the proposed continuous distance field is
compared with the state-of-the-art continuous GP-based approach
by~\citet{le_gentil_accurate_2024}. The accuracy measures for both the distance
and gradient prediction are detailed in~\cref{ssec:perf-meas}. For the proposed
collision probability calculation method, the improvement due to the
blending approach (\cref{eq:gmm-prob}) is analyzed for different levels of noise in robot position.
Experiments are conducted using real-world and simulated 3D
(\cref{ssec:3d-eval}) point clouds.

\subsection{Computational Performance\label{ssec:perf-compute}}
For this analysis, $100000$ pairs of random ellipsoids are created by
randomizing axis lengths, positions, and rotations of 3D ellipsoids.  Axis
lengths and positions (in all directions) are sampled uniformly from the
intervals $[0.1, 0.5]$\SI{}{\meter} and $[-10, 10]$\SI{}{\meter}, respectively.
Random rotation matrices are generated using the Haar
distribution~\citep{stewart_efficient_1980}. The time taken for ellipsoid pair
data structure initialization, distance and gradient computation, and collision
probability calculation is measured. The initialization procedure caches
matrices $\hat{\mathbf{B}}, \boldsymbol{\Lambda}_B, \mathbf{B}^{1/2},
\mathbf{B}^{-1/2}$, and $\mathbf{B}^{-1}$ for
$\ellipsoid{}{1}{\mathbf{b}}{\mathbf{B}}$ (and the respective counterparts for
$\ellipsoid{}{2}{\mathbf{c}}{\mathbf{C}}$), as defined in~\cref{lem:dist}.
Distance and gradient computation times correspond to
calculating~\cref{eq:ell-dist}.  For collision probability
calculation,~\cref{eq:ell-prob} is used.  The performance is measured on one
desktop (Intel i9-10900K) and three embedded platforms (NVIDIA Orin AGX, Orin
NX, and Orin Nano).

The mean and standard deviation statistics of elapsed time (in microseconds) are
reported in~\cref{tab:compute}. It can be concluded that the time taken by all
computers in~\cref{tab:compute} is between $10$ to $60$ microseconds.
The distance and gradient computation times are shown in a single column because
computing the gradient direction requires computing the distance vector.
The total time (last column) represents the worst case where new ellipsoids are
initialized every time the distance, gradient, and collision probability are
calculated. In practice, the ellipsoids are initialized and updated using a few
sensor observations; not for every distance or collision probability
computation. As expected, the computation time increases as the clock speed of
the CPU decreases.

Since real-world motion planning systems largely run on CPUs, these observations
indicate that the proposed methods may be beneficial for adapting many existing
motion planners towards leveraging GSMs as the environment representation.
Furthermore, since the proposed methods run efficiently on a CPU, the GPU can be
used for tasks like online high-fidelity RGB-D reconstruction. Fast
fitting of GSMs to point cloud data is an active area of research, with recent
approaches demonstrating high frame-rate operation on embedded computers like
the NVIDIA TX2~\citep{li_gmmap_2024}.

\begin{table}
  \centering
  \caption{Mean and standard deviation of the time taken (in microseconds)
  to initialize a pair of ellipsoids and estimate distance, gradient, and
  collision probability using single-threaded execution on various CPUs.}\label{tab:compute}
  {
  \footnotesize
  \begin{tabular}{@{}lcccc@{}}
  \toprule
  & \multicolumn{4}{c}{Time Taken ($10^{-6}\SI{}{\second}$), single-threaded execution}\\
  \cmidrule{2-5}
  Device (CPU) & Init.  & Dist. + Grad. & Coll. Prob. & Total\\
  \cmidrule{1-1} \cmidrule(lr){2-2} \cmidrule(lr){3-3} \cmidrule(lr){4-4} \cmidrule(lr){5-5}
  i9 \SI{3.7}{\giga\hertz} & $9.0 \pm 1.7$ & $9.5 \pm 2.0$ & $2.2 \pm 0.9$ & $20.7 \pm 3.0$\\
  AGX \SI{2.2}{\giga\hertz} & $18.4 \pm 1.1$ & $17.3 \pm 1.8$ & $5.3 \pm 0.6$ & $41.0 \pm 2.3$ \\
  NX \SI{2.0}{\giga\hertz} & $22.4 \pm 4.2$ & $22.9 \pm 2.6$ & $6.7 \pm 0.8$ & $52.0 \pm 5.2$ \\
  Nano \SI{1.5}{\giga\hertz} & $24.5 \pm 1.6$ & $24.8 \pm 2.7$ & $7.0 \pm 0.8$ & $56.2 \pm 3.4$ \\
\end{tabular}
  }
  \vspace{-2em}
\end{table}

\subsection{Accuracy Measures\label{ssec:perf-meas}}
The accuracy in the distance field is measured (in meters) using the
Root-Mean-Squared-Error (RMSE) between the prediction and the ground truth. A
lower RMSE indicates higher accuracy. For the gradient of the distance
field, we use $1.0 - \text{RMSE}(\cos(\mathbf{D}_p, \mathbf{D}_{g}))$, where the
cosine is computed at each corresponding point in the predicted ($\mathbf{D}_p$)
and the ground truth vector field ($\mathbf{D}_g$)~\citep[Eq. 15]{ortiz_isdf_2022}.
This score is referred to as the Cosine Error Score (CES) in the following
sections. A lower CES implies a better alignment between the predicted and the
ground truth gradient.

The ground truth distance between ellipsoid and point cloud is computed using
the point cloud distance between points densely sampled on the robot ellipsoid
surface and the surface point cloud.
The ground truth gradient vector is
given by the surface normal at the point in the surface point cloud that is
closest to the robot point cloud.

\begin{figure*}
  \begin{minipage}{0.5\columnwidth}
  \centering
  \ifthenelse{\equal{\arxivmode}{false}}%
  {%
  \includegraphics[width=\columnwidth,trim=550 200 600 200, clip]{figures/pcld_3d_real.eps}\\
  \subfloat[Point Cloud\label{sfig:pcld_3d_real}]{\includegraphics[width=\columnwidth,trim=400 600 700 50, clip]{figures/copyroom_pcld.eps}}%
  }%
  {%
  \includegraphics[width=\columnwidth,trim=550 200 600 200, clip]{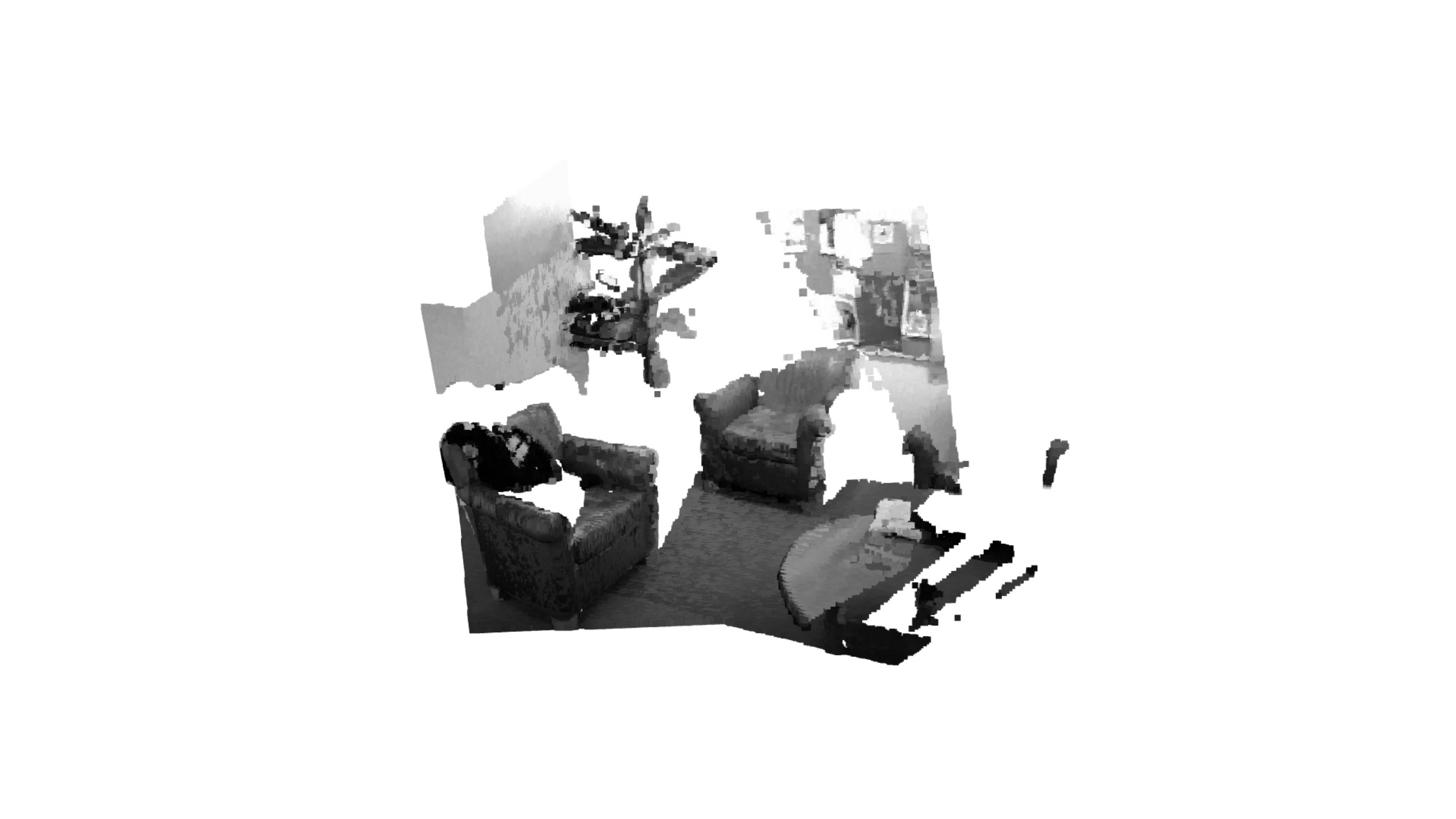}\\
  \subfloat[Point Cloud\label{sfig:pcld_3d_real}]{\includegraphics[width=\columnwidth,trim=400 600 700 50, clip]{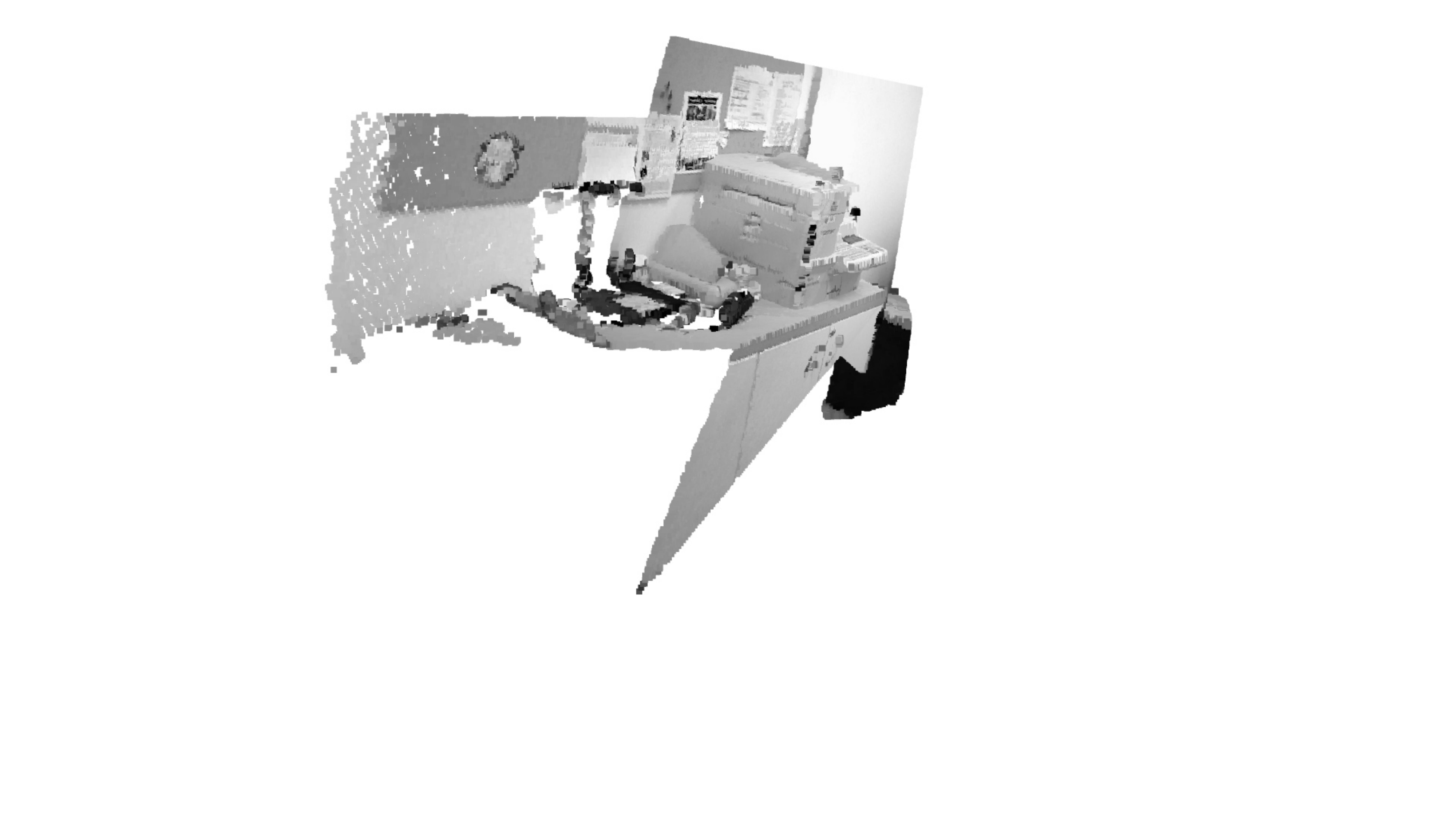}}%
  }%
  \end{minipage}%
  \begin{minipage}{0.5\columnwidth}
  \centering
  \ifthenelse{\equal{\arxivmode}{false}}%
  {%
  \includegraphics[width=\columnwidth,trim=300 100 500 200, clip]{figures/gt_df_3d_real.eps}\\
  \subfloat[Ground Truth\label{sfig:gt_df_3d_real}]{\includegraphics[width=\columnwidth,trim=300 220 400 50,clip]{figures/copyroom_gt_df.eps}}%
  }%
  {%
  \includegraphics[width=\columnwidth,trim=300 100 500 200, clip]{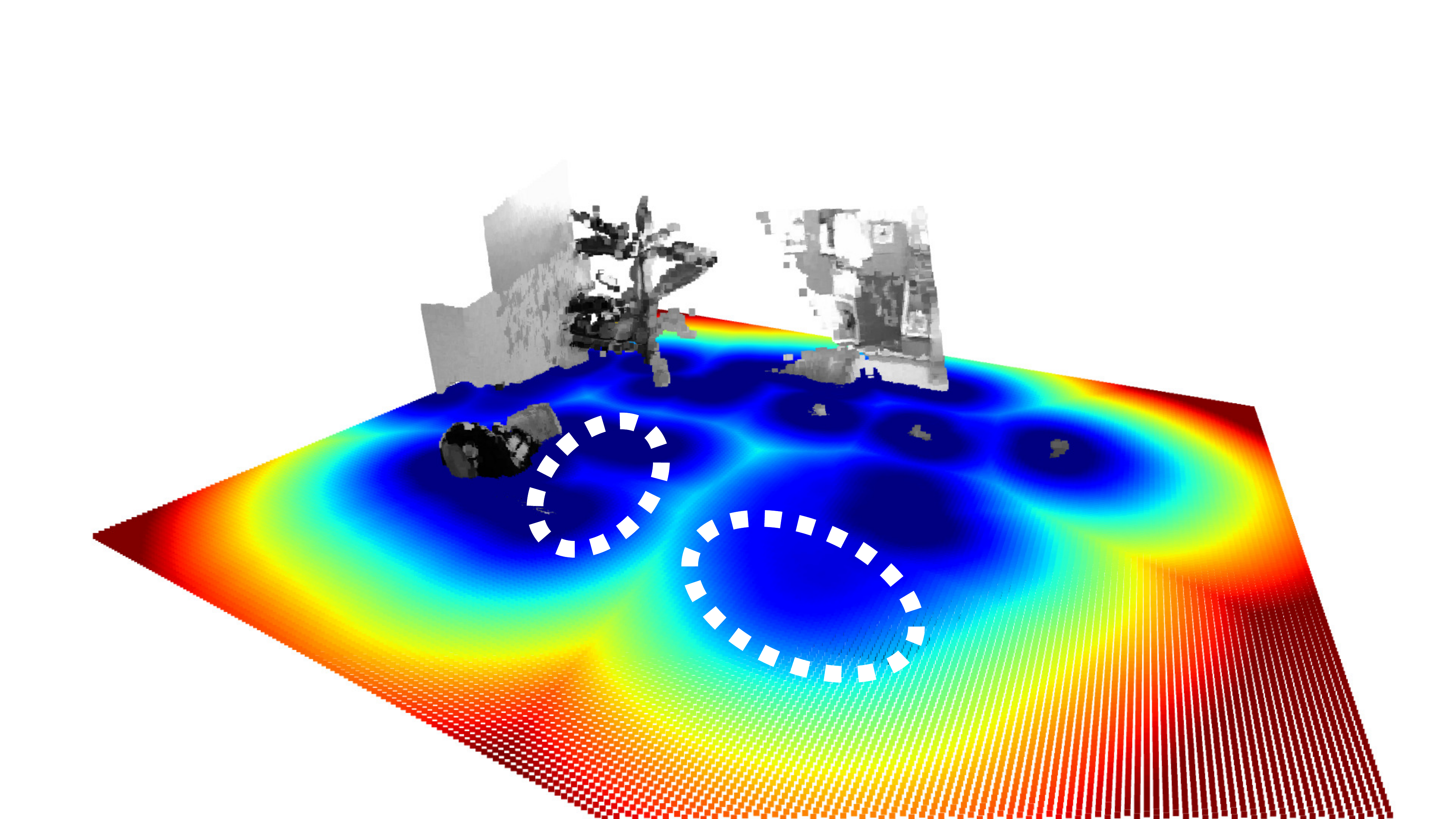}\\
  \subfloat[Ground Truth\label{sfig:gt_df_3d_real}]{\includegraphics[width=\columnwidth,trim=300 220 400 50,clip]{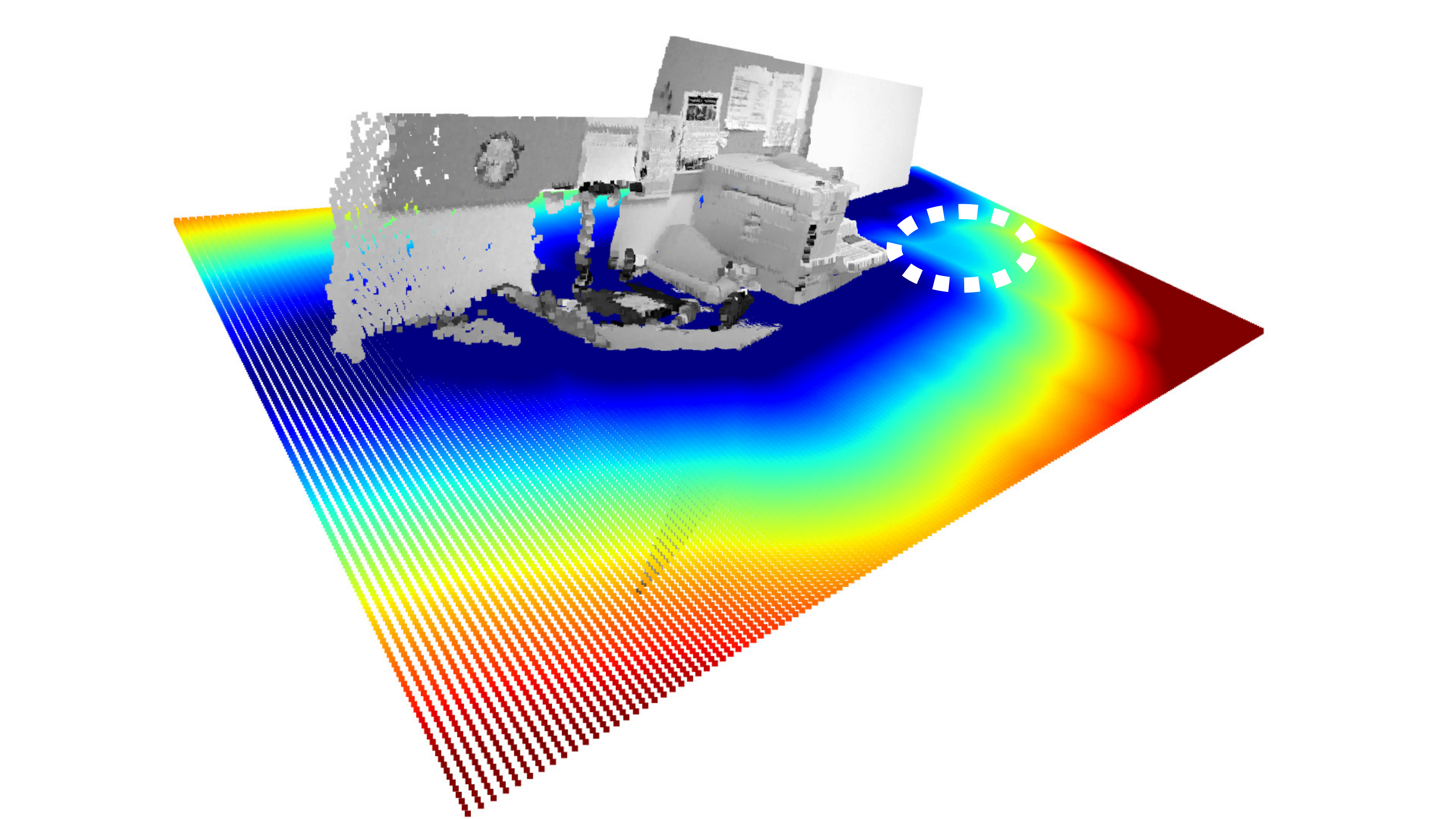}}%
  }%
  \end{minipage}
  \begin{minipage}{0.5\columnwidth}
  \centering
  \ifthenelse{\equal{\arxivmode}{false}}%
  {%
  \includegraphics[width=\columnwidth,trim=300 100 500 200, clip]{figures/se_df_3d_real.eps}\\
  \subfloat[Baseline~\citep{le_gentil_accurate_2024} EDF\label{sfig:se_df_3d_real}]{\includegraphics[width=\columnwidth,trim=300 220 400 50,clip]{figures/copyroom_se_df.eps}}%
  }%
  {%
  \includegraphics[width=\columnwidth,trim=300 100 500 200, clip]{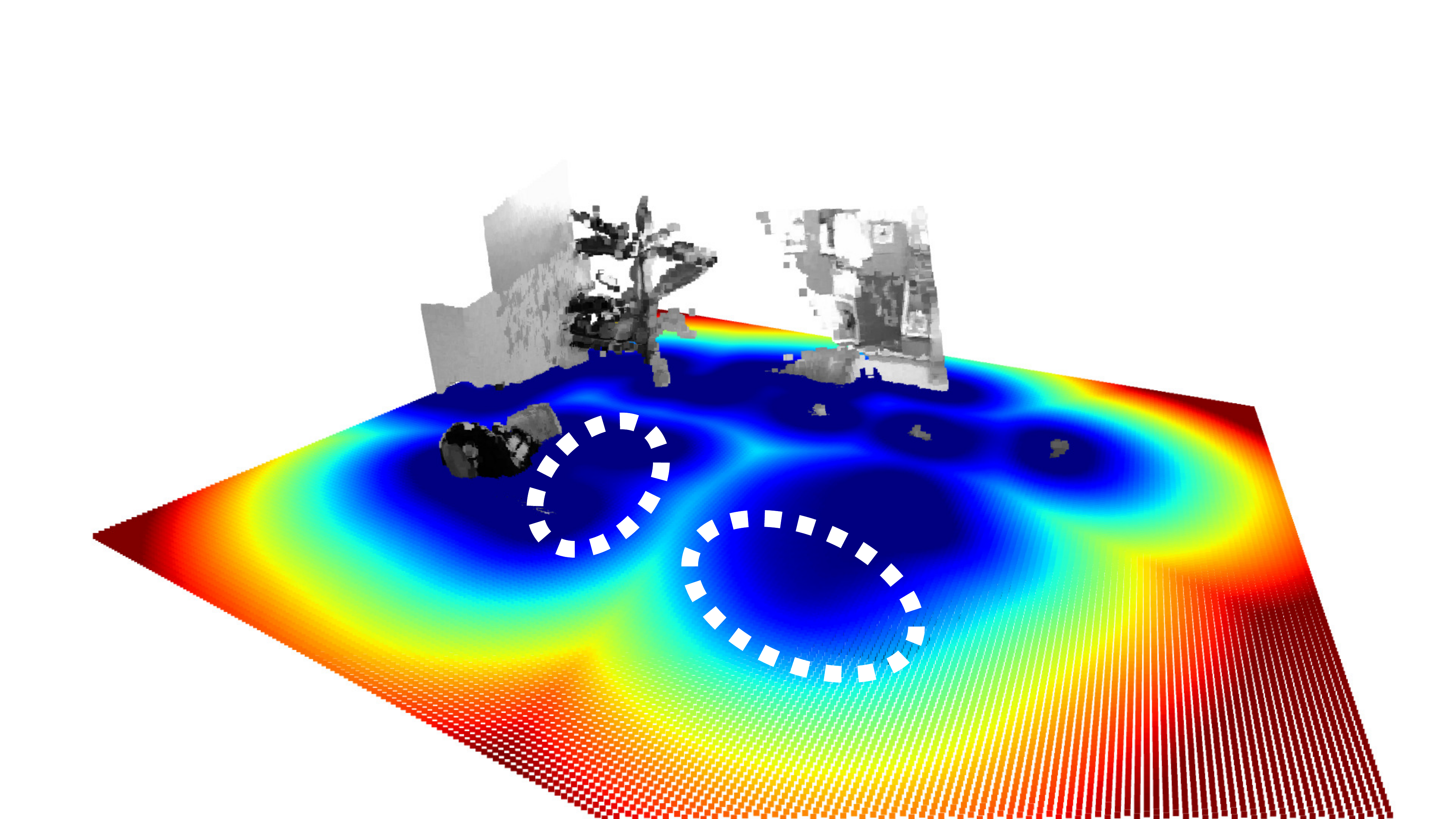}\\
  \subfloat[Baseline~\citep{le_gentil_accurate_2024} EDF\label{sfig:se_df_3d_real}]{\includegraphics[width=\columnwidth,trim=300 220 400 50,clip]{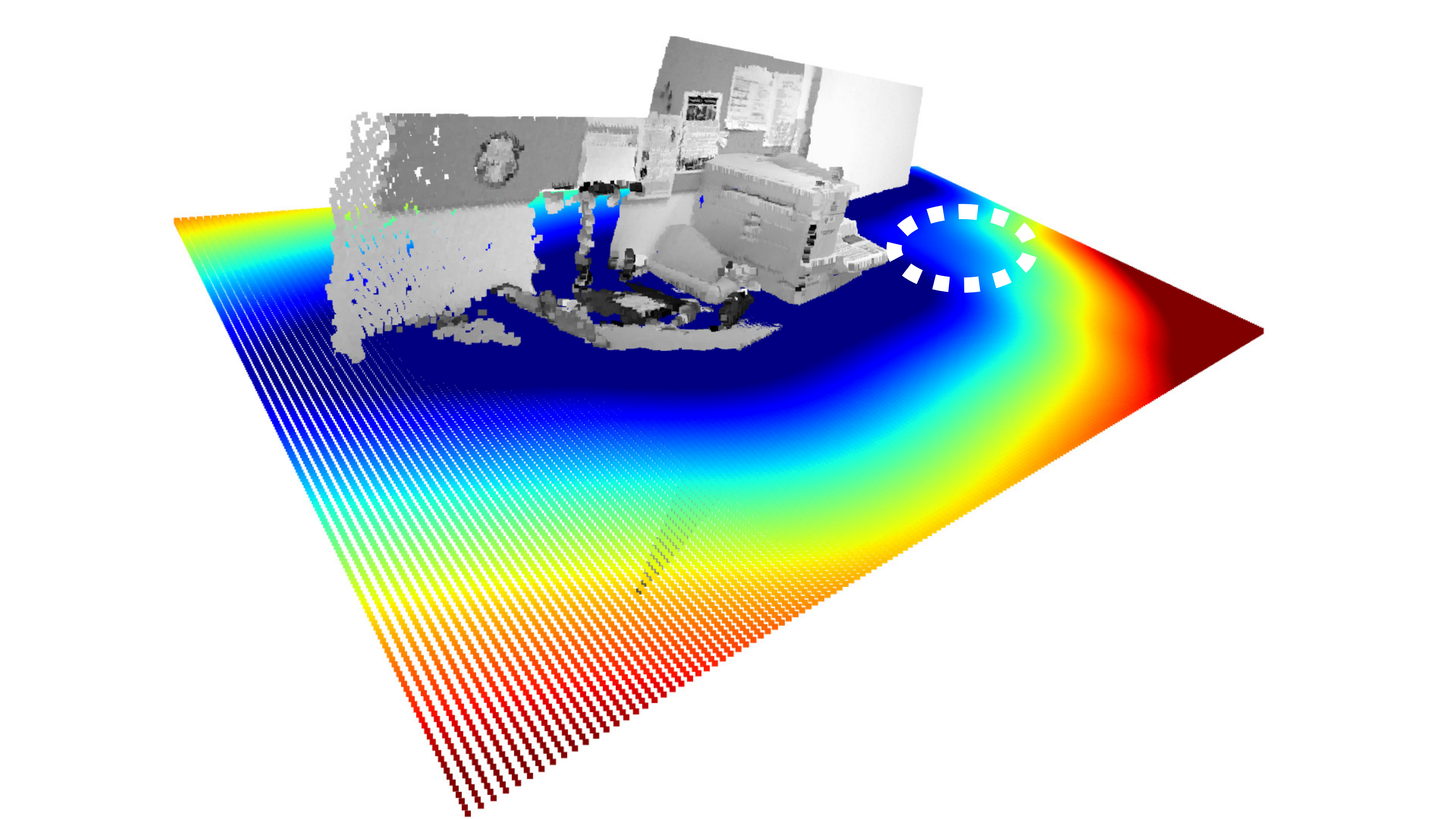}}%
  }%
  \end{minipage}
  \begin{minipage}{0.5\columnwidth}
  \centering
  \ifthenelse{\equal{\arxivmode}{false}}%
  {%
  \includegraphics[width=\columnwidth,trim=300 100 500 200, clip]{figures/df_3d_real.eps}\\
  \subfloat[Proposed EDF\label{sfig:prop_df_3d_real}]{\includegraphics[width=\columnwidth,trim=300 220 400 50,clip]{figures/copyroom_prop_df.eps}}%
  }%
  {%
  \includegraphics[width=\columnwidth,trim=300 100 500 200, clip]{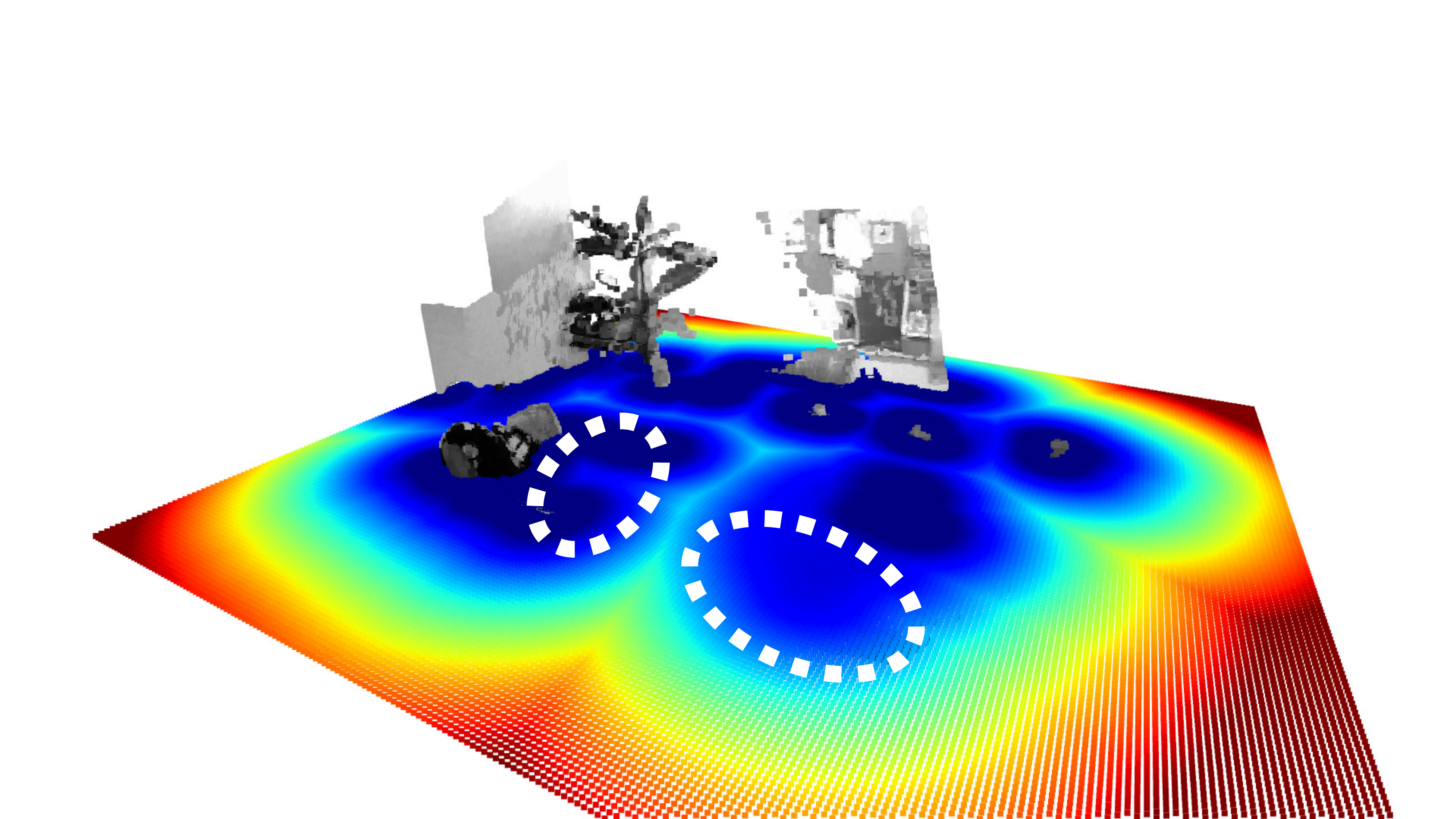}\\
  \subfloat[Proposed EDF\label{sfig:prop_df_3d_real}]{\includegraphics[width=\columnwidth,trim=300 220 400 50,clip]{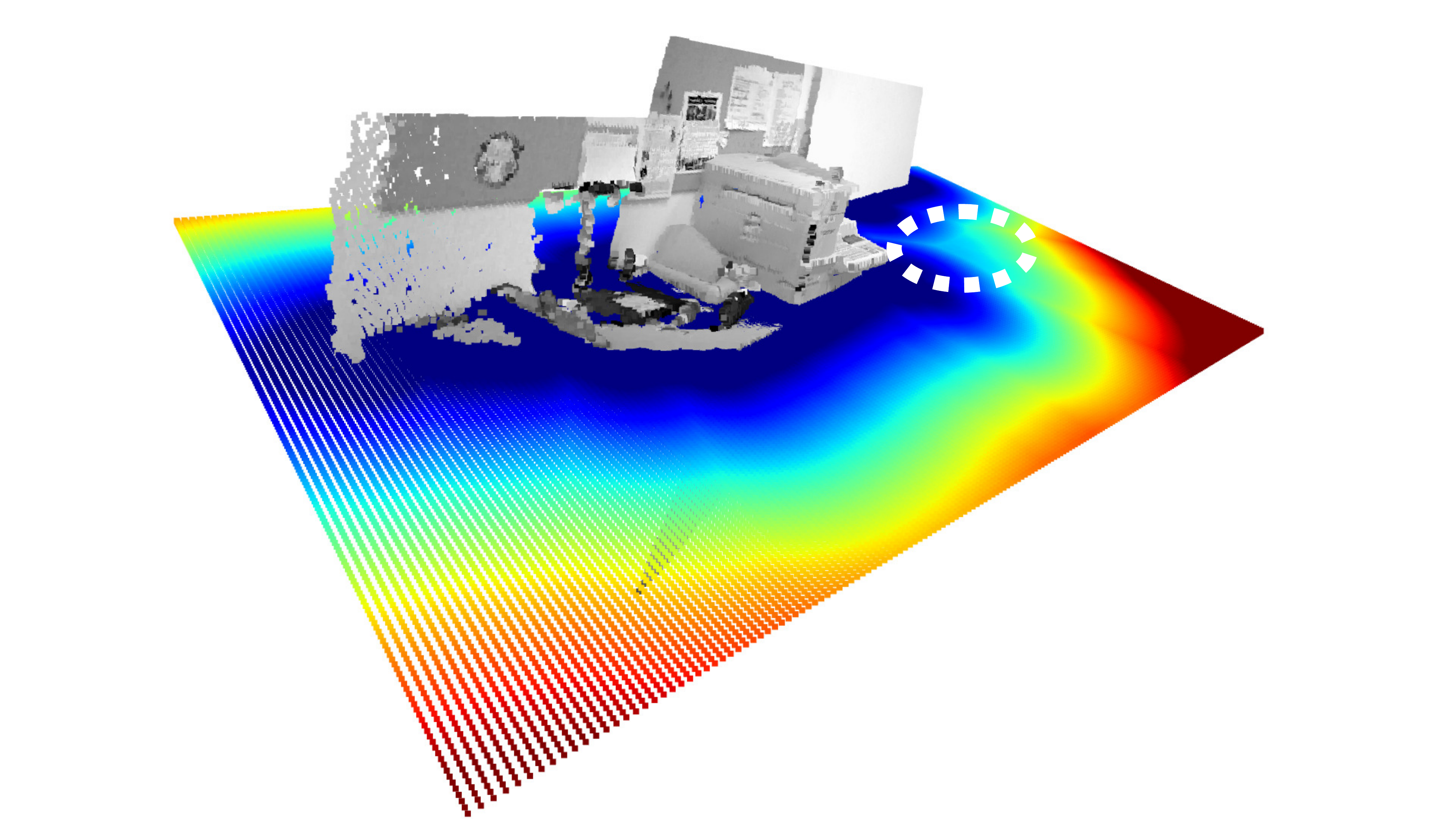}}%
  }%
  \end{minipage}
  \caption{\label{fig:edf-3d-real} Heatmaps for
  \protect\subref{sfig:gt_df_3d_real} ground truth,
  \protect\subref{sfig:se_df_3d_real} baseline~\citep{le_gentil_accurate_2024},
  and~\protect\subref{sfig:prop_df_3d_real} proposed EDFs generated using the real-world 3D point cloud shown
  in~\protect\subref{sfig:pcld_3d_real}. Note the difference in baseline and proposed EDFs relative to the
  ground truth in the dashed white regions. The dark blue regions are bigger in
  the baseline demonstrating conservative EDF estimation due to an implicit
  spherical robot body assumption. The proposed approach accounts for the
  ellipsoidal robot body while enabling continuous-space queries.
  }
\end{figure*}

\begin{figure}
  \centering
  \begin{minipage}{\columnwidth}
    \centering
    \ifthenelse{\equal{\arxivmode}{false}}%
    {%
    \input{figures/colorbar-fig8.tex}\\
    \subfloat[\citep{le_gentil_accurate_2024}, Simulated]{\input{figures/se_err_livingroom1.tex}}%
    \subfloat[\citep{le_gentil_accurate_2024}, Real \textit{Lounge}]{\input{figures/se_err_lounge.tex}}%
    \subfloat[\citep{le_gentil_accurate_2024}, Real \textit{Copyroom}]{\input{figures/se_err_copyroom.tex}}%
    }%
    {%
    \includegraphics{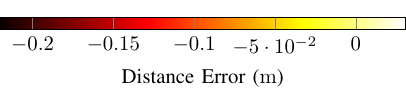}\\
    \subfloat[\citep{le_gentil_accurate_2024}, Simulated]{\includegraphics{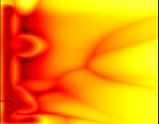}}\hfil
    \subfloat[\citep{le_gentil_accurate_2024}, Real \textit{Lounge}]{\includegraphics{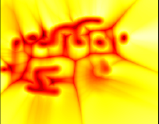}}\hfil
    \subfloat[\citep{le_gentil_accurate_2024}, Real \textit{Copyroom}]{\includegraphics{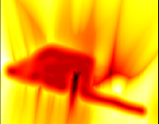}}%
    }%
  \end{minipage}\\
  \begin{minipage}{\columnwidth}
    \centering
    \ifthenelse{\equal{\arxivmode}{false}}%
    {%
    \subfloat[Ours, Simulated]{\input{figures/prop_err_livingroom1.tex}}%
    \subfloat[Ours, Real \textit{Lounge}]{\input{figures/prop_err_lounge.tex}}%
    \subfloat[Ours, Real \textit{Copyroom}]{\input{figures/prop_err_copyroom.tex}}
    }%
    {%
    \subfloat[Ours, Simulated]{\includegraphics{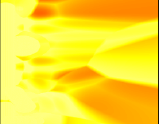}}\hfil
    \subfloat[Ours, Real \textit{Lounge}]{\includegraphics{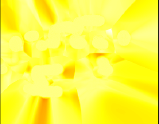}}\hfil
    \subfloat[Ours, Real \textit{Copyroom}]{\includegraphics{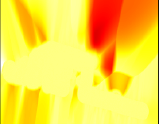}}
    }%
  \end{minipage}
  \caption{\label{fig:3d-df-err}Distance field estimation error heatmaps
  for the baseline and proposed approaches. The proposed approach enables
  a lower estimation error compared to the baseline.
  }
\end{figure}

\begin{figure}
  \centering
  \begin{minipage}{\columnwidth}
  \centering
  \ifthenelse{\equal{\arxivmode}{false}}%
  {%
  \input{figures/colorbar-fig10.tex}\\
  \subfloat[Unblended, Sim., $\sigma^2_p = 0.01$\label{sfig:ubl-3d-0-01}]{\input{figures/ubl_3d_0.01_livingroom.tex}}%
  \subfloat[Blended, Sim., $\sigma^2_p = 0.01$\label{sfig:bl-3d-0-01}]{\input{figures/bl_3d_0.01_livingroom.tex}}%
  }%
  {%
  \includegraphics{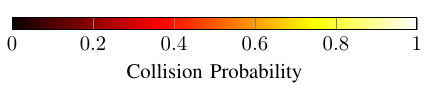}\\
  \subfloat[Unblended, Sim., $\sigma^2_p = 0.01$\label{sfig:ubl-3d-0-01}]{\includegraphics{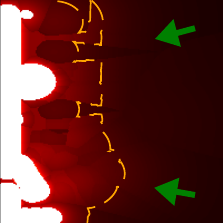}}\hfil
  \subfloat[Blended, Sim., $\sigma^2_p = 0.01$\label{sfig:bl-3d-0-01}]{\includegraphics{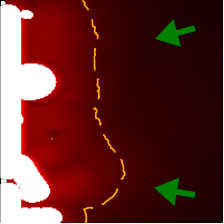}}%
  }%
  \end{minipage}\\
  \begin{minipage}{\columnwidth}
  \centering
  \ifthenelse{\equal{\arxivmode}{false}}%
  {%
  \subfloat[Unblended, Real, $\sigma^2_p = 0.04$\label{sfig:ubl-lounge-0-04}]{\input{figures/ubl_3d_0.04_lounge.tex}}%
  \subfloat[Blended, Real, $\sigma^2_p = 0.04$\label{sfig:bl-lounge-0-04}]{\input{figures/bl_3d_0.04_lounge.tex}}%
  }%
  {%
  \subfloat[Unblended, Real, $\sigma^2_p = 0.04$\label{sfig:ubl-lounge-0-04}]{\includegraphics{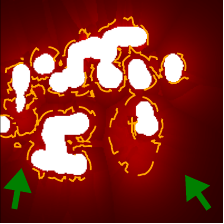}}\hfil
  \subfloat[Blended, Real, $\sigma^2_p = 0.04$\label{sfig:bl-lounge-0-04}]{\includegraphics{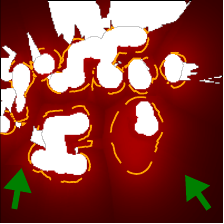}}%
  }%
  \end{minipage}\\
  \begin{minipage}{\columnwidth}
  \centering
  \ifthenelse{\equal{\arxivmode}{false}}%
  {%
  \subfloat[Unblended, Real, $\sigma^2_p = 0.04$\label{sfig:ubl-copyroom-0-04}]{\input{figures/ubl_3d_0.04_copyroom.tex}}%
  \subfloat[Blended, Real, $\sigma^2_p = 0.04$\label{sfig:bl-copyroom-0-04}]{\input{figures/bl_3d_0.04_copyroom.tex}}%
  }%
  {%
  \subfloat[Unblended, Real, $\sigma^2_p = 0.04$\label{sfig:ubl-copyroom-0-04}]{\includegraphics{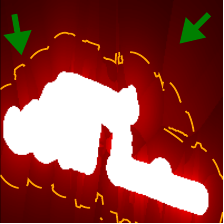}}\hfil
  \subfloat[Blended, Real, $\sigma^2_p = 0.04$\label{sfig:bl-copyroom-0-04}]{\includegraphics{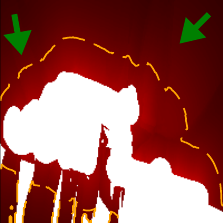}}%
  }%
  \end{minipage}
  \caption{\label{fig:3d-coll-prob}Unblended and blended collision probability fields over 2D slices of 3D
  simulated and real-world point clouds. More noise in robot position is added
  to the real-world cases to simulate the effect of higher position uncertainty
  during real-world deployments. Dashed lines show $10\%$ probability
  isocontours. \textcolor{green}{Green} arrows show the directions of camera frustums from which
  the surface point cloud data is collected. The blending approach produces smoother collision probability
  estimates while ignoring occluded regions.
  }
\end{figure}

\begin{table}
  \centering
  \caption{\label{tab:3d-quant-edf}Quantitative results using
  simulated and real 3D point clouds at different hyperparameter settings for
  the baseline and proposed methods. The best RMSE and CES values for each
  dataset are bolded.
  }
  {
  \footnotesize
  \addtolength{\tabcolsep}{-0.4em}
\begin{tabular}{@{}lcccccc@{}}
  \toprule
  & \multicolumn{2}{c}{Simulated} & \multicolumn{2}{c}{\textit{Lounge}} & \multicolumn{2}{c}{\textit{Copyroom}}\\
  \cmidrule(lr){2-3} \cmidrule(lr){4-5} \cmidrule(lr){6-7}
  Approach & RMSE $\downarrow$ &  CES $\downarrow$ & RMSE $\downarrow$ &  CES $\downarrow$ & RMSE $\downarrow$ &  CES $\downarrow$\\
  \citep{le_gentil_accurate_2024} ($0.1$) & 0.054 & 0.166 & 0.039 & 0.258 & 0.056 & \textbf{0.277}\\
  \citep{le_gentil_accurate_2024} ($0.2$) & 0.084 & 0.168 & 0.061 & 0.258 & 0.075 & 0.278\\
  \citep{le_gentil_accurate_2024} ($0.3$) & 0.119 & 0.173 & 0.098 & 0.258 & 0.100 & 0.283\\
  \citep{le_gentil_accurate_2024} ($0.4$) & 0.145 & 0.171 & 0.132 & \textbf{0.257} & 0.129 & 0.290\\
  \midrule
  Ours ($0.02$) & 0.042  & \textbf{0.159}  & \textbf{0.023}  & 0.263 & \textbf{0.046} & 0.289\\
  Ours ($0.03$) & \textbf{0.038}  & 0.164  & 0.042  & 0.263 & 0.058 & 0.295\\        
  Ours ($0.04$) & 0.068  & 0.181  & 0.038  & 0.267 & 0.057 & 0.305\\                
  Ours ($0.05$) & 0.085  & 0.191  & 0.054  & 0.267 & 0.062 & 0.296\\
  \bottomrule 
\end{tabular}
  }
  \vspace{-2em}
\end{table}

\subsection{3D Experiments\label{ssec:3d-eval}}
Three point clouds are used in 3D experiments: a simulated point cloud from the
Living Room dataset~\citep{choi_robust_2015}, a real-world point cloud from the
Lounge dataset~\citep{zhou_dense_2013}, and a real-world point cloud from the
Copyroom dataset~\citep{zhou_dense_2013}. The point clouds are constructed using
two $320 \times 240$ RGB-D frames from each dataset
(see~\cref{sfig:pcld_3d,sfig:pcld_3d_real}). The accuracy of Euclidean distance
field (EDF) and its gradient is studied for 2D uniform grid ($200 \times 200$)
slices of these environments, as done in prior
work~\citep{wu_log-gpis-mop_2023,le_gentil_accurate_2024}.  The robot is a 3D
ellipsoid with semi-axis lengths $(0.15, 0.15, 0.07)$ and it is rotated about
the $z$-axis by $45$ degrees.~\Cref{sfig:prop_df_3d,sfig:prop_bl_3d} show EDF
and collision probability outputs for the proposed approach on the simulated
point cloud.  We empirically set $K = 9$ in~\cref{eq:gmm-prob}. 

\Cref{fig:edf-3d-real} shows a qualitative comparison of the EDF obtained from
different methods for the real point clouds. The proposed approach is relatively
accurate because it accounts for the ellipsoidal robot body explicitly. For a
better visualization of the difference in errors incurred by the two
methods,~\cref{fig:3d-df-err} contains error heatmaps for the EDFs
in~\cref{fig:edf-3d-real}.

For different values of hyperparameters the quantitative comparison of EDFs and
gradient is summarized in~\cref{tab:3d-quant-edf}. The range of hyperparameter
$l$ for the baseline approach is chosen by grid search to avoid numerical
instability. For $\sigma$, the range of values are based on the results
in~\citep{goel_probabilistic_2023}. It is observed that for both simulated and
real-world cases, the proposed approach enables relatively accurate EDFs at all
hyperparameter levels while providing a comparable gradient accuracy.

Let the center of the robot ellipsoid be Gaussian-distributed with the spherical
covariance $\sigma^2_p \mathbf{I}_3$, where $\mathbf{I}_n$ denotes the identity
matrix of order $n$. The unblended and blended collision probabilities for
$\sigma^2_p = 0.01$ and $0.04$ along with isocontours for probability level
$0.15$ are shown in~\cref{fig:3d-coll-prob}. It is observed that for increasing
levels of noise there is an overall increase in collision probabilities.
Moreover, in the occluded regions (i.e., regions behind the surface when viewed
along the green arrows in~\cref{fig:3d-coll-prob}) the blended approach does not
provide reliable estimates. This is because the dot product between the surface
normals and distance vectors is negative. It is reasonable to expect the
estimates in the occluded regions to be degraded until the occluded regions are
observed.  The unblended approach provides estimates without this consideration
which may be risky during navigation.  The blending approach results in smoother
probability isocontour lines (lower noise in dashed orange lines) in the visible
of the 3D experiments, also, making the approach valuable for continuous-space
queries in future 3D uncertainty-aware motion planning frameworks.


\section{Conclusion}\label{sec:conclusion}
This work detailed collision probability, Euclidean distance and gradient
estimation for an ellipsoidal robot from a surface that is represented as a set
of ellipsoids derived from Gaussian distributions. Prior work in
ellipsoid-to-ellipsoid distance estimation was extended to compute distance and
gradient in the proposed context. A geometrical blending approach ensured that
the estimated collision probabilities are smooth so that they can be used for
uncertainty-aware motion planning. These methods were validated using 3D
real-world point cloud environments, demonstrating superior performance compared
to the state-of-the-art continuous space method. Lastly, the computational
performance of the proposed methods indicates that the distance, gradient, and
collision probability estimation can be executed in a single-thread on low-power
CPUs of embedded computers (e.g. NVIDIA Orin) in a few milliseconds for
thousands of ellipsoids. In future work, we will integrate the
approach with an incremental GMM mapping strategy
(e.g.,~\citep{dhawale_efficient_2020,goel_incremental_2023,li_gmmap_2024})
for robot navigation in large environments.

There are two key limitations of this work. First, the computation
in~\cref{eq:gmm-dist} may require additional local submap extraction or spatial
partitioning data structures to enable scalability as $M$
increases. A concurrent or vectorized implementation of the eigenvalue problems
in~\cref{lem:dist} may further improve performance. Second, estimation
in the orientation space (formally, in the special orthogonal group
$\mathbb{SO}(2)$ or $\mathbb{SO}(3)$) is not explicitly considered in this work
which may be an interesting direction for future research.


\section*{Acknowledgments}
This work was supported in part by an Uber Presidential Fellowship. This
material is based upon work supported by, or in part by, the Army Research
Laboratory and the Army Research Office under contract/grant number
W911NF-25-2-0153. The authors thank J.~Lee, M.~Hansen, and D.~Wettergreen for
feedback on this manuscript.

{
  \balance
  \footnotesize
  \bibliographystyle{plainnatlink}
  \bibliography{refs,content/bibliography/kshitij_library}
}

\end{document}